%% file: main.tex
\documentclass[10pt,twocolumn,letterpaper]{article}
  \providecommand{\llamaindexbranded}{}
  \usepackage[margin=1in]{geometry}
\usepackage[utf8]{inputenc}
\usepackage[T1]{fontenc}
\usepackage{times}
\usepackage{amsmath,amssymb}
\usepackage{graphicx}
\graphicspath{{figures/}}
\usepackage{subcaption}
\usepackage{capt-of}
\usepackage{booktabs}
\usepackage{tabularx}
\newcolumntype{L}[1]{>{\raggedright\arraybackslash\hspace{0pt}}X}
\usepackage{colortbl}
\usepackage{longtable}
\usepackage{multirow}
\usepackage{makecell}
\usepackage{enumitem}
\usepackage{numprint}
\npstyleenglish
\usepackage{etoc}
\usepackage{float}
\usepackage{placeins}
\usepackage{tikz}
\usetikzlibrary{positioning,arrows.meta,calc,fit,backgrounds}

\usepackage{xcolor}
\ifdefined\cleanbuildenabled
\newcommand{\boyang}[1]{}
\newcommand{\simon}[1]{}
\newcommand{\justin}[1]{}
\newcommand{\todo}[1]{}
\else
\newcommand{\boyang}[1]{{\color{red}\textbf{[Boyang:} #1\textbf{]}}}
\newcommand{\simon}[1]{{\color{blue}\textbf{[Simon:} #1\textbf{]}}}
\newcommand{\justin}[1]{{\color{teal}\textbf{[Justin:} #1\textbf{]}}}
\newcommand{\todo}[1]{{\color{orange}\textbf{[TODO:} #1\textbf{]}}}
\fi

\newcommand{\cmark}{$\checkmark$}

\definecolor{oursband}{HTML}{E8F1FB}

  \usepackage[colorlinks=true,linkcolor=blue,citecolor=blue,urlcolor=blue]{hyperref}
\usepackage[nameinlink,capitalize]{cleveref}
\ifdefined\llamaindexbranded
  \usepackage{llamapaper}
\fi

\usepackage[numbers,sort&compress]{natbib}
\usepackage{xfp}
\newcommand{\statnum}[1]{\numprint{#1}}
\input{sections/stats_macros}

\input{sections/result_macros}

\title{ExtractBench Pro: The Long Tail of Document Extraction}

\author{
  Boyang Zhang$^{*}$ \quad
  Justin Xu$^{*}$ \quad
  Adrian Lyjak \quad
  Eli Stewart \quad
  Simon Suo \\[4pt]
  \normalsize \texttt{\{boyang, justin, adrian, eli, simon\}@runllama.ai}
}
\date{}
\hypersetup{pdfauthor={Boyang Zhang, Justin Xu, Adrian Lyjak, Eli Stewart, Simon Suo}}

\ifdefined\llamaindexbranded
\LIPaperType{Technical report}
\LIPaperDate{November 2026}
\LIPDFAuthor{Boyang Zhang, Justin Xu, Adrian Lyjak, Eli Stewart, Simon Suo}
\LIPDFSubject{A benchmark for document extraction when the value is not printed on the page}
\LIAbstract{\input{sections/abstract}}
\LIPaperLinks{%
  \href{https://huggingface.co/datasets/llamaindex/ExtractBenchPro}{Dataset}
  \hspace{1.1em}
  \href{https://github.com/run-llama/ExtractBench}{Evaluation code}
  \hspace{1.1em}
  \href{https://www.extractbench.ai/pro/}{Website}
}
\fi

\begin{document}

\maketitle
{\renewcommand{\thefootnote}{\fnsymbol{footnote}}\footnotetext[1]{Equal contribution.}}
\ifdefined\llamaindexbranded\else
\begin{abstract}
\input{sections/abstract}
\end{abstract}
\fi

\begin{figure*}[t]
\centering
\input{figures/page_samples}
\caption{Selected page regions from ten generated ExtractBench Pro documents spanning all six challenge groups, illustrating their visual diversity and complexity.}
\label{fig:hero}
\end{figure*}

\input{sections/01_intro}
\input{sections/02_related_work}
\input{sections/03_extractbench_pro}
\input{sections/04_experiments}
\input{sections/05_conclusion}

\clearpage
\begingroup
\small
  \bibliographystyle{plainnat}
\bibliography{references}
\endgroup

\clearpage
\onecolumn
\input{sections/appendix}

\end{document}

%% file: sections/stats_macros.tex
\newcommand{\prodocs}{100}
\newcommand{\propipelines}{90}
\newcommand{\propages}{2622}

\newcommand{\provalues}{144141}

\newcommand{\proboxedshare}{78}
\newcommand{\proboxeddocs}{81}

\newcommand{\pronsystems}{18}
\newcommand{\proPerDocSystems}{18}
\newcommand{\proUnionPerfect}{91}

\newcommand{\prondomains}{7}

%% file: sections/result_macros.tex
\newcommand{\amPerfect}{56}
\newcommand{\amValue}{92.7}

\newcommand{\amCostLong}{\numprint{332}}

\newcommand{\dlCostLong}{40.5}

\newcommand{\axPerfect}{59}

\newcommand{\opPerfectThirtyMinute}{51}
\newcommand{\opValueThirtyMinute}{75.4}

\newcommand{\opPerfect}{61}
\newcommand{\opValue}{93.1}

\newcommand{\bestAgentGroundedLongCanon}{27.7}
\newcommand{\bestApiGroundedLongCanon}{0.0}
\newcommand{\longSheetsCostMaxCents}{\numprint{1139}}
\newcommand{\otherChallengesCostMaxCents}{19.9}
\newcommand{\agentLongSheetsCostMinCents}{11.1}
\newcommand{\bestPerfectName}{Claude Code Opus 5.5}
\newcommand{\bestPerfect}{61}
\newcommand{\bestPerfectCost}{41.2}

\newcommand{\axLongMaxCostUsd}{20}

\newcommand{\bottomTenLongSheets}{4}
\newcommand{\bottomTenWithPerfect}{6}
\newcommand{\bottomTenMinBest}{96}

\newcommand{\topThreeUnionPerfect}{84}
\newcommand{\costFrontierNames}{LlamaExtract Cost Effective, LlamaExtract Agentic, Datalab Accurate, LlamaExtract Agentic Plus Max Effort, and Claude Code Opus 5.5}

%% file: sections/abstract.tex
We present ExtractBench Pro, a schema-guided document extraction benchmark that tests six challenging modes of extraction underrepresented in existing datasets. It is created following a framework for synthetic document generation, in which deterministic document generators assemble challenging components into documents that are designed to be difficult at a level calibrated for frontier systems. We synthesize 100 realistic documents spanning a diverse range of styles. On this data, we evaluate ten coding agents and eight extraction API configurations for accuracy, grounding, and their ability to perfectly extract documents. No extraction system extracts more than \bestPerfect{}\% of documents error-free. The strongest general-purpose coding agents, Claude Code Opus 5.5 and Codex GPT-6 Astra, outperform specialized extraction APIs on both perfect-document rate and Value F1, but at a higher cost per page. Among specialized extraction APIs, LlamaExtract spans the lower-cost frontier, with Agentic Plus Max Effort approaching the strongest coding agent on both perfect-document rate and Value F1 at lower cost per page. Our analysis of coding agent traces identifies intelligent image processing and programmatic reasoning, but also failures in applying document-specific definitions and recovering relationships from diagrams. We release ExtractBench Pro and these evaluation results as resources for the future development of improved frontier extraction systems.

%% file: figures/page_samples.tex
\begingroup
\begin{minipage}[t]{0.19\textwidth}\centering\includegraphics[width=\linewidth]{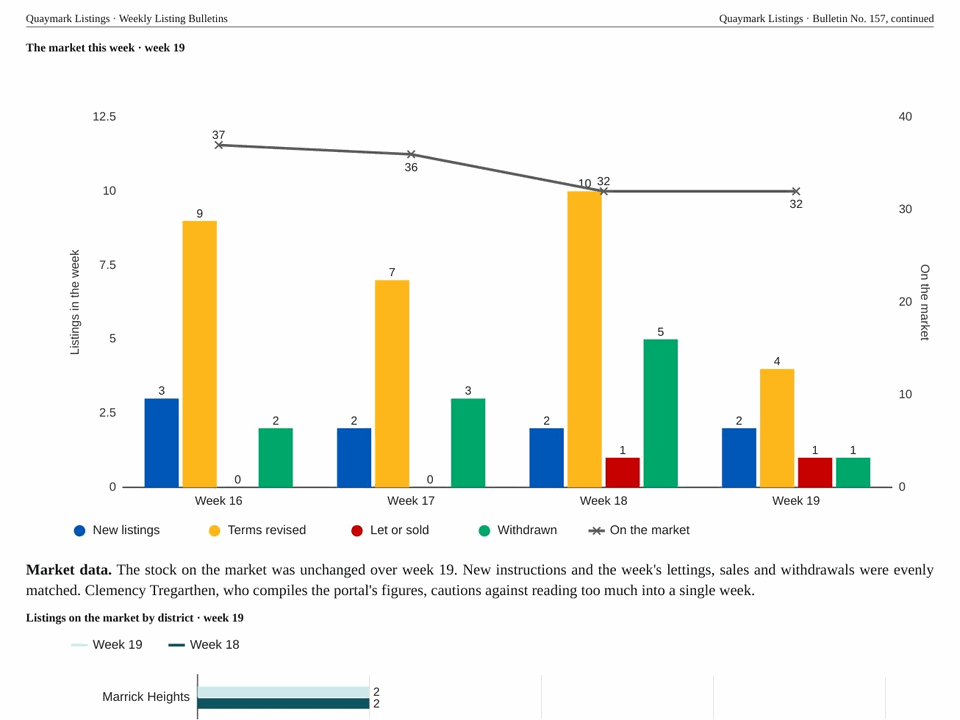}\\[2pt]{\scriptsize\textbf{Interpretation rules}\\(a) Property market bulletin}\end{minipage}\hfill
\begin{minipage}[t]{0.19\textwidth}\centering\includegraphics[width=\linewidth]{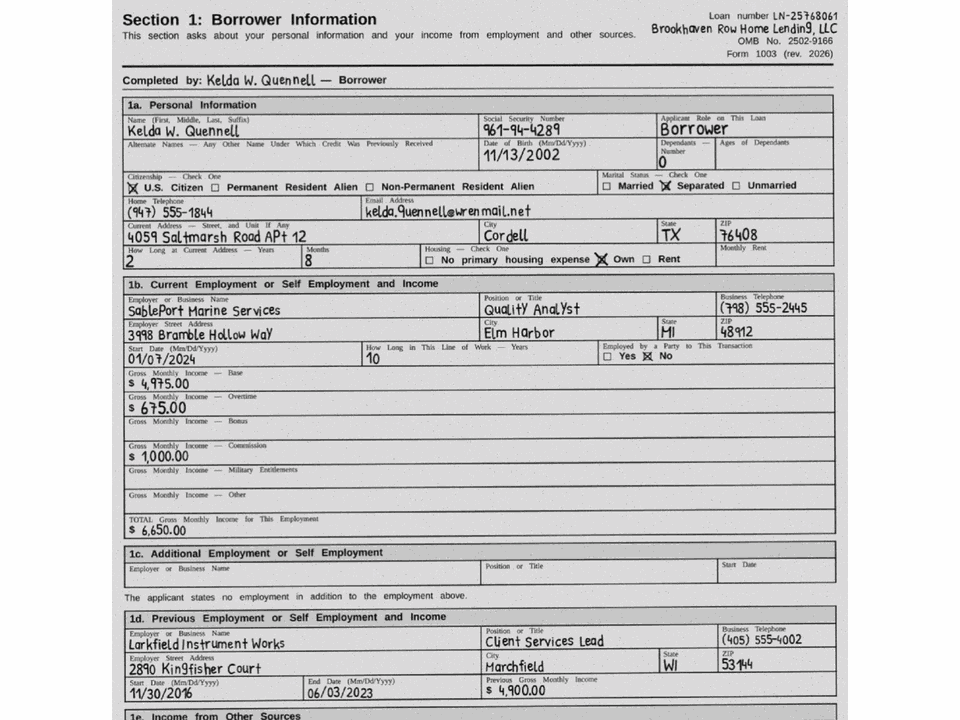}\\[2pt]{\scriptsize\textbf{Interpretation rules}\\(b) Mortgage application}\end{minipage}\hfill
\begin{minipage}[t]{0.19\textwidth}\centering\includegraphics[width=\linewidth]{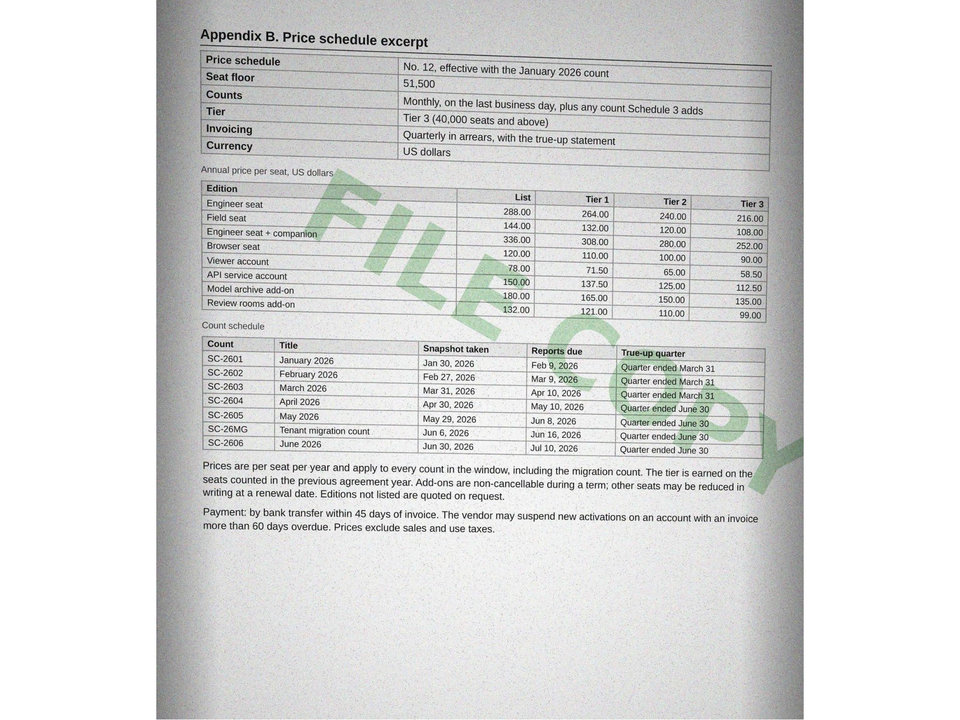}\\[2pt]{\scriptsize\textbf{Custom terminology}\\(c) Subscription-seat review}\end{minipage}\hfill
\begin{minipage}[t]{0.19\textwidth}\centering\includegraphics[width=\linewidth]{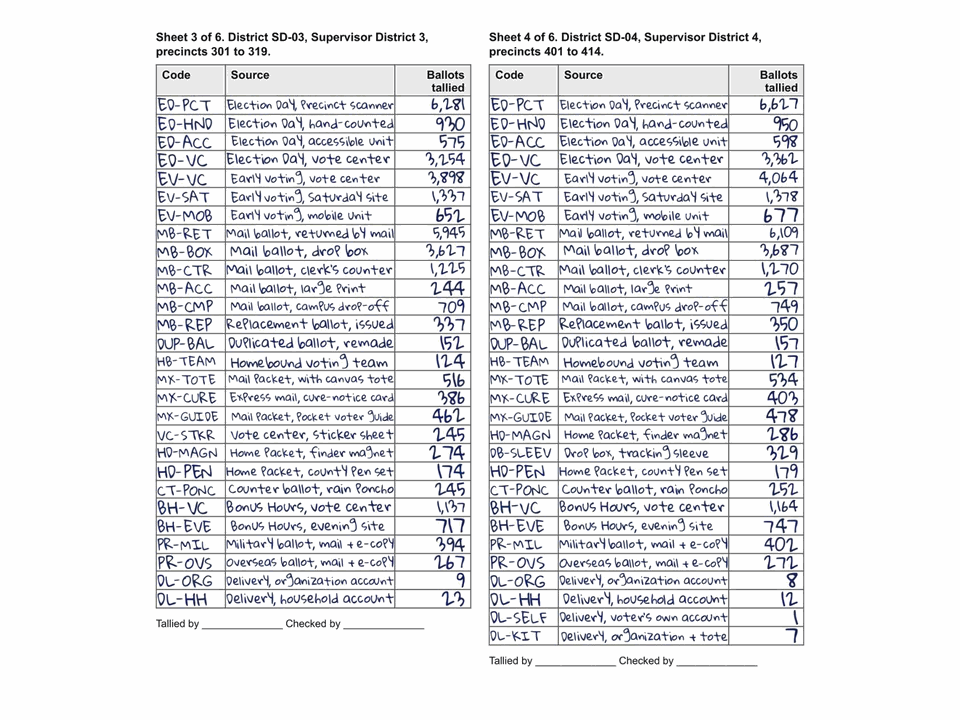}\\[2pt]{\scriptsize\textbf{Conditional aggregation}\\(d) Election canvass binder}\end{minipage}\hfill
\begin{minipage}[t]{0.19\textwidth}\centering\includegraphics[width=\linewidth]{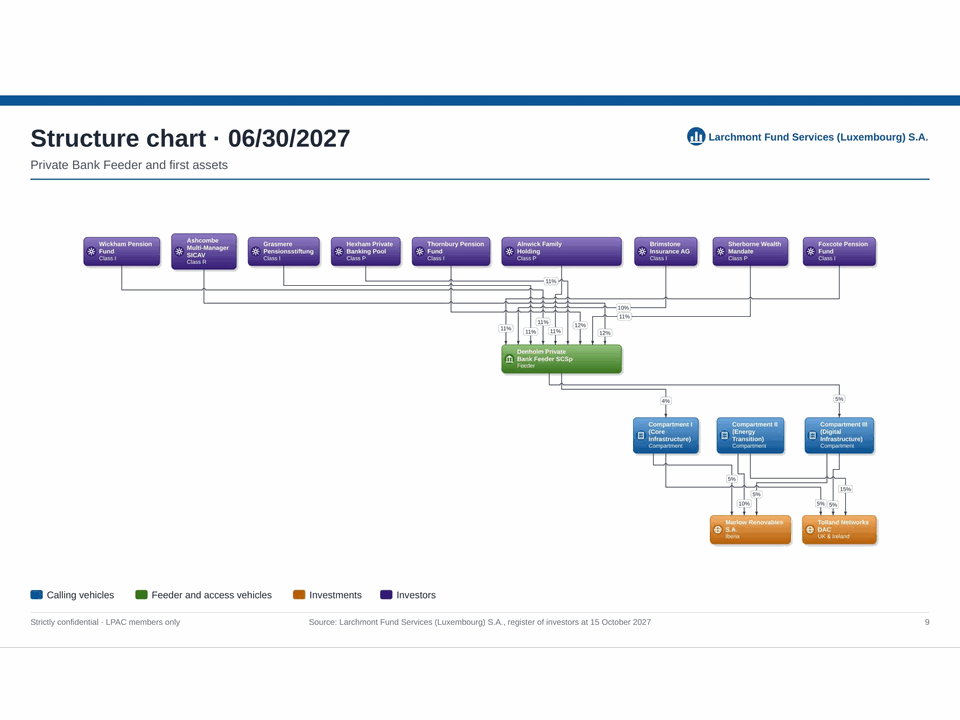}\\[2pt]{\scriptsize\textbf{Relational diagrams}\\(e) Fund structure chart}\end{minipage}
\par\vspace{8pt}
\begin{minipage}[t]{0.19\textwidth}\centering\includegraphics[width=\linewidth]{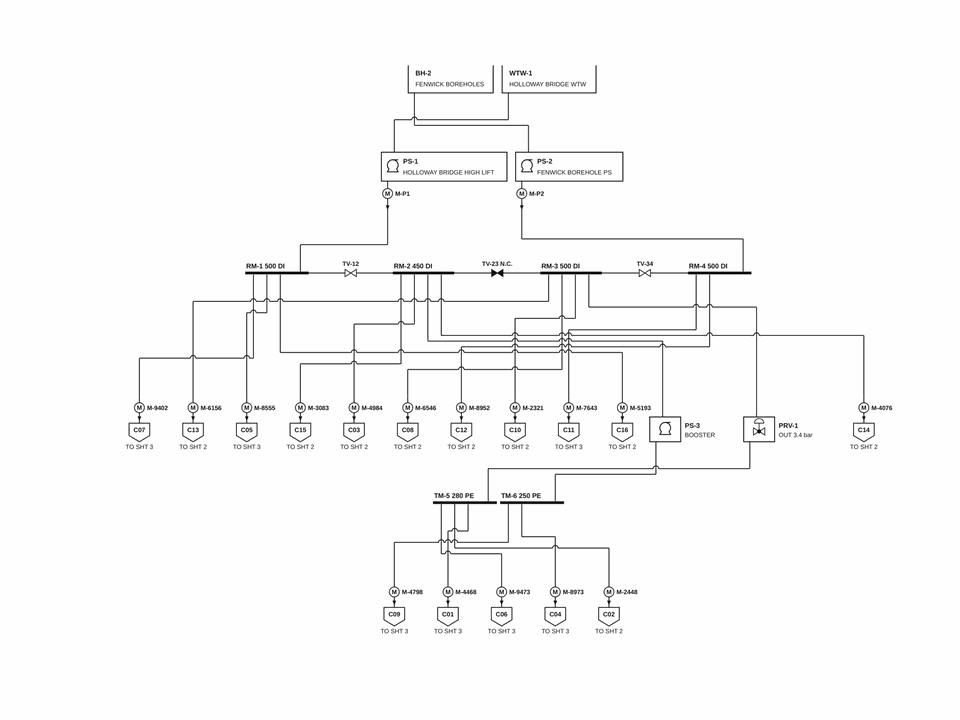}\\[2pt]{\scriptsize\textbf{Relational diagrams}\\(f) Water network audit}\end{minipage}\hfill
\begin{minipage}[t]{0.19\textwidth}\centering\includegraphics[width=\linewidth]{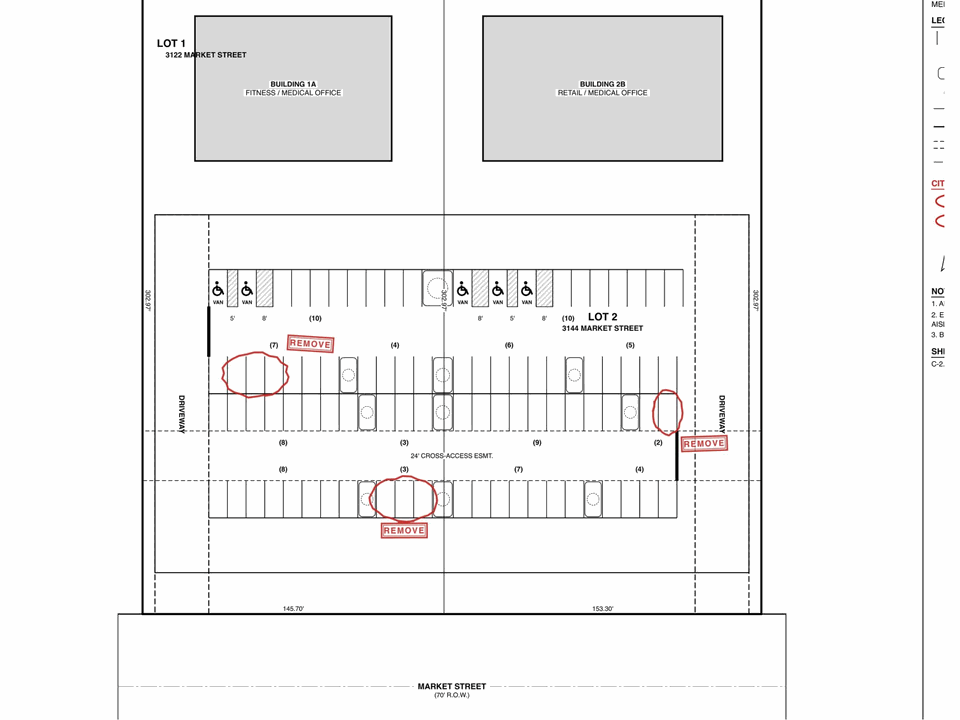}\\[2pt]{\scriptsize\textbf{Charts and maps}\\(g) Site plan with review marks}\end{minipage}\hfill
\begin{minipage}[t]{0.19\textwidth}\centering\includegraphics[width=\linewidth]{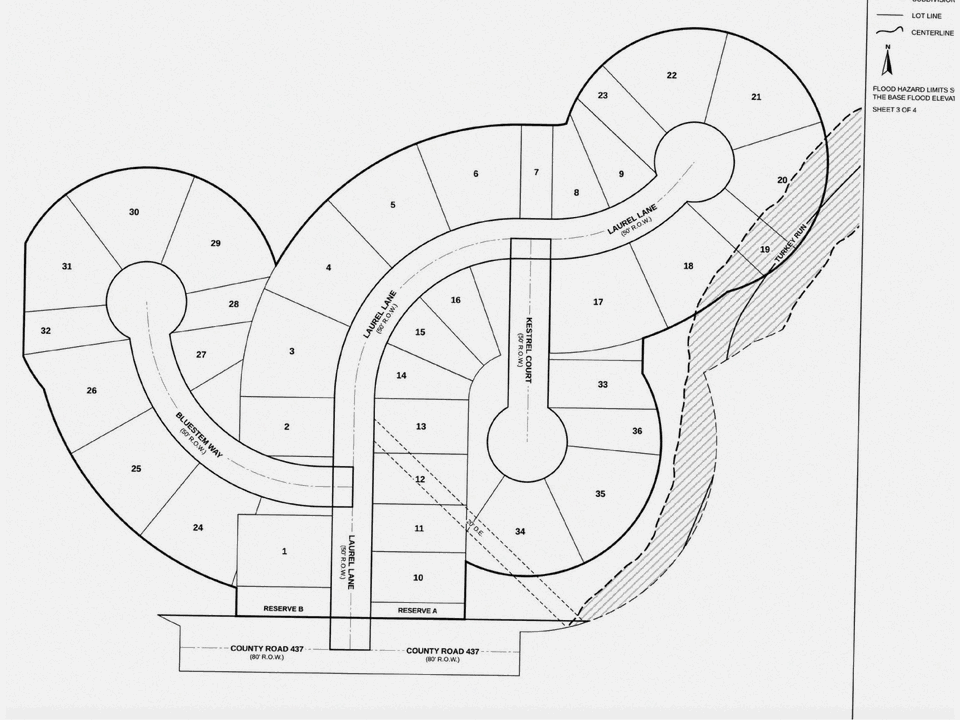}\\[2pt]{\scriptsize\textbf{Charts and maps}\\(h) Subdivision plat}\end{minipage}\hfill
\begin{minipage}[t]{0.19\textwidth}\centering\includegraphics[width=\linewidth]{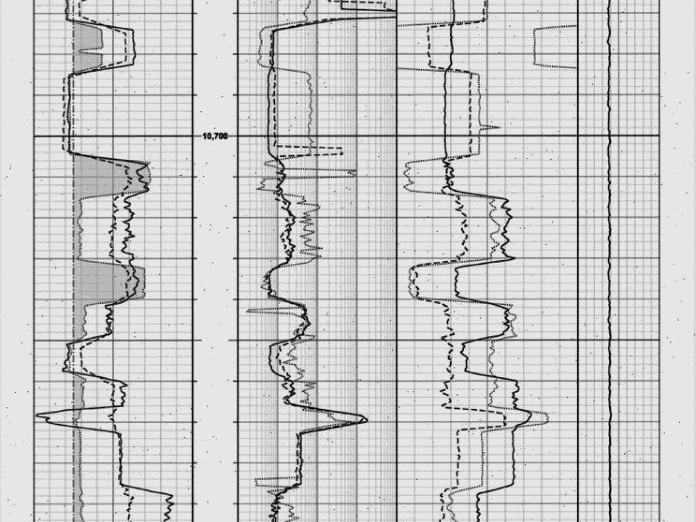}\\[2pt]{\scriptsize\textbf{Long single pages}\\(i) Wireline log 1}\end{minipage}\hfill
\begin{minipage}[t]{0.19\textwidth}\centering\includegraphics[width=\linewidth]{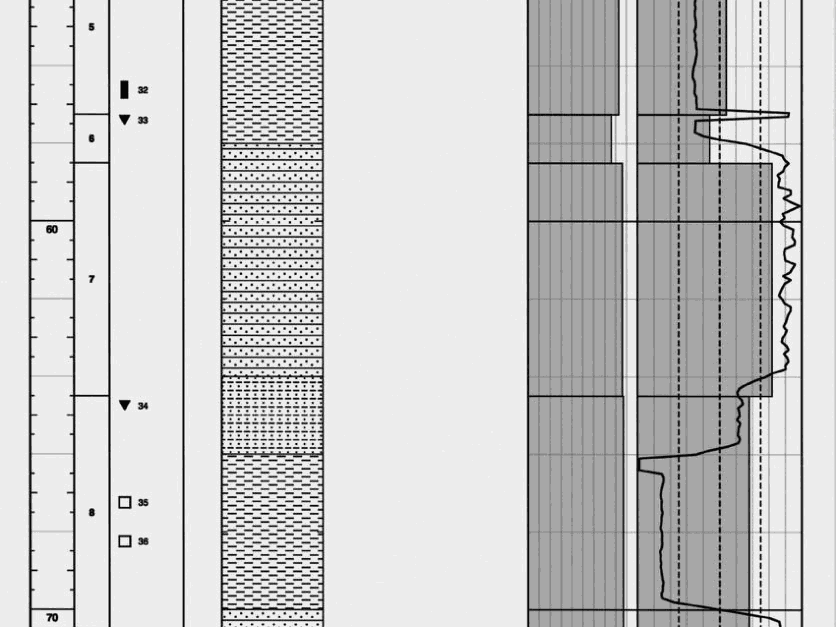}\\[2pt]{\scriptsize\textbf{Long single pages}\\(j) Borehole core log}\end{minipage}
\endgroup

%% file: sections/01_intro.tex
\section{Introduction}
\label{sec:intro}

An initial task in many automated document processing workflows is to convert an unstructured PDF document into a structured JSON object, while following a schema. It enables valuable automation in domains such as finance, legal, and manufacturing, on documents such as tax forms, case filings, and spec sheets. However, the standard of quality needed for many of these applications is perfection. Mistakes incur costly additional human review and correction, especially if they are not caught until later in a workflow. Therefore, accuracy and price, in that order, tend to be the most important metrics to measure in the task of schema-guided document extraction.

Accuracy is measured, sometimes alongside cost, on benchmarks such as ExtractBench~\cite{zhang2026extractbench}, LongArray-Extract~\cite{extend2026longarray}, and VAREX~\cite{varex2026}. Each benchmark typically targets a set of challenges representative of a specific set of real-world extraction workloads. However, many extractors now rank above 95\% on the headline metrics of most existing benchmarks~\cite{extend2026longarrayresults,extractbench2026leaderboard}, driven by advances in vision language models (VLMs) and their harnesses. Despite this, many real-world documents still suffer poor extraction performance, even with the most powerful models. Domains such as engineering and planning feature documents containing non-textual elements like relational diagrams, charts, and maps. Private documents also tend to have custom terminology and interpretation rules that change the meaning of even textual elements. This constitutes a long tail of document extraction, and a gap in existing benchmarks that makes it difficult for practitioners to assess which extractor is best suited to real-world applications. A benchmark is needed that can controllably target challenges in the long-tail gap, at a difficulty level appropriate for evaluating frontier systems.

In this work, we present ExtractBench Pro, a benchmark for the long tail of document extraction. To meet frontier benchmarking needs, it is constructed following a framework for synthetic document generation, using coding agents to write deterministic generators that produce documents, schemas, and ground truth extractions from simple inputs. We refer to real-world long-tail documents to design a set of reusable components that can be instantiated into elements which are challenging to extract. These elements are composed to form documents, with precise control over style and difficulty. The resulting dataset consists of 100 documents split into six challenge groups: interpretation rules, custom terminology, conditional aggregation, relational diagrams, charts and maps, and long single pages.

We evaluate specialized extraction APIs and coding agents on ExtractBench Pro. Extraction APIs tend to have lower, usually fixed, costs per page compared to coding agents which achieve generally higher accuracy through iterative reasoning and code execution. Nevertheless, the benchmark is difficult for all extractors. The highest-performing system by perfect-document rate, \bestPerfectName{}, extracts just \bestPerfect{}\% of documents perfectly, at a cost of \bestPerfectCost{} \textcent{}/page. Among specialized APIs, LlamaExtract's tiers trace the lower end of the cost-quality frontier. LlamaExtract's highest-accuracy offering, Agentic Plus Max Effort, extracts 56 documents perfectly at 36.6\textcent{} per page, compared with 59 perfect documents by Codex GPT-6 Astra at more than 2.5 times the cost per page. We analyze the traces of coding agents and discover impressive instances of image processing and programmatic reasoning in the most successful extractors, providing intuitions for how frontier systems may further advance performance on ExtractBench Pro.

Our contributions are:

\begin{itemize}[leftmargin=1.2em,itemsep=2pt]

\item \textbf{A configurable framework for generating documents and ground truth together.} Generators deterministically assemble high-fidelity documents modelled after real-life examples. They can be calibrated to challenge frontier extraction systems.

\item \textbf{A challenging benchmark for schema-guided document extraction.} ExtractBench Pro covers seven business domains and six challenge groups requiring document rules, calculations, and visual interpretation of non-textual elements.

\item \textbf{A systematic evaluation of frontier extraction systems.} We compare ten coding agents and eight extraction API configurations on Value F1, perfect extraction rate, Grounding F1, cost, and completion time. We examine their traces and mistakes and develop intuitions for how future extraction systems may improve.

\end{itemize}

%% file: sections/02_related_work.tex
\input{tables/benchmark_comparison_options}

\section{Related Work}
\label{sec:related-work}

\paragraph{Methods for Document Extraction.}

Single-pass extraction involves a general-purpose or specialized VLM reading the document and generating its output without iterative inspection. Context and output limits are challenges for long tasks~\cite{ferguson2026extractbench,micro1longextract} such as those found in this work, excluding these approaches from our evaluation.

Coding agents combine a VLM with a general-purpose harness and tools. Applying those tools toward document extraction, an agent can inspect documents repeatedly, compute intermediate results, access files, and revise its outputs. Codex and Claude Code accept the document and schema as files and can use code to navigate pages, process records, and check a candidate answer~\cite{anthropic2026claudecode,openai2026codex}. These operations are relevant when evidence is distributed, or requested values require calculations. However, repeated inspection and computation increase cost and may delay completion.

Meanwhile, specialized extractors provide managed workflows for document processing and structured output. Services such as Datalab, Extend, LlamaExtract, and Reducto handle document ingestion and schema-guided extraction through an API; some also return source locations~\cite{datalab2026extract,extend2026operator,llamaindex2026llamaextract,reducto2026deep}. We evaluate these services alongside coding agents to measure how their different approaches handle the same combinations of reasoning, visual interpretation, and long-document navigation.

\paragraph{Benchmarks for Document Extraction.}

\label{sec:related-extractbench}

Early document-extraction benchmarks evaluate a single predefined set of fields. FUNSD and SROIE study forms and receipts~\cite{huang2019sroie,jaume2019funsd}; Kleister extends extraction to long documents~\cite{stanislawek2021kleister}, while DocILE and RealKIE add business layouts, line items, and enterprise domains~\cite{simsa2023docile,townsend2025realkienoveldatasetsenterprise}.

Schema-guided benchmarks test whether systems can adapt to a set of fields provided at extraction time. Contextual AI's ExtractBench emphasizes schema breadth and nested outputs~\cite{ferguson2026extractbench}, ExStrucTiny examines reformulated requests and answer localization~\cite{sibue2026exstructiny}, and VAREX and UniKIE-Bench vary schemas, layouts, and document conditions~\cite{varex2026,ji2026unikie}. LongArray-Extract tests recovery of complete record arrays as their length, formatting, and page organization vary~\cite{extend2026longarray}. LongExtractionBench evaluates completeness and execution failures on long reports and filings with large, repeated tables~\cite{micro1longextract}.

The main basis of this work, ExtractBench, extends schema-guided evaluation across varied extraction tasks and document conditions~\cite{zhang2026extractbench}. It introduces the simultaneous evaluation of systems along value accuracy, source grounding, and cost. \Cref{tab:benchmark-comparison} compares their challenge coverage and evaluation measures. ExtractBench Pro builds on these evaluation axes to target the long tail of extraction tasks that remain difficult for frontier systems. We use controlled synthesis to calibrate task difficulty while preserving realism and diversity.

\paragraph{Synthetic Document Generation.}

Synthetic documents support evaluation by pairing known ground truth with controllable content and presentation. They let one construct rare or difficult cases that are hard to collect, and vary selected properties while holding others fixed to isolate failure causes, as VAREX does with document representations~\cite{varex2026}. The challenge is to achieve this control while capturing the diversity and fidelity of real-world documents.

Prior work has progressed from layouts mined from real documents to configurable components and schemas, as in DocCreator and FlexDoc~\cite{dua2025flexdoc,journet2017doccreator,raman2022synthetic}. Achieving realism through these approaches depends on increasingly sophisticated construction rules. DocGenie and DocDjinn use reference images to guide VLM-generated content and layout, with DocDjinn also producing task annotations~\cite{harikrishnan2025docgenie,lamott2026docdjinn}. This offers flexibility, but generative models within document generation can introduce hallucinations and reduce reproducibility~\cite{harikrishnan2025docgenie}. Our approach uses coding agents to make sophisticated generation programs practical to implement and refine, grounded in real documents. The resulting programs jointly produce documents, schemas, and annotations through deterministic execution, with configurable content and presentation that let us compose difficult tasks for frontier extraction systems.

%% file: tables/benchmark_comparison_options.tex
\begin{table*}[t!]
\centering
\footnotesize
\setlength{\tabcolsep}{2.5pt}
\begin{tabular*}{\textwidth}{@{\extracolsep{\fill}}lccccccc@{}}
\toprule
 & \multicolumn{5}{c}{\textbf{Extraction challenges}} & \multicolumn{2}{c}{\textbf{Evaluation}} \\
\cmidrule(lr){2-6}\cmidrule(lr){7-8}
\textbf{Benchmark} & \makecell{\textbf{Long}\\\textbf{lists}} & \makecell{\textbf{Degraded}\\\textbf{pages}} & \makecell{\textbf{Cross-}\\\textbf{references}} & \makecell{\textbf{Computed}\\\textbf{values}} & \makecell{\textbf{Chart}\\\textbf{evidence}} & \textbf{Grounding} & \textbf{Cost} \\
\midrule
Contextual ExtractBench~\cite{ferguson2026extractbench} & \cmark & & & & & & \\
ExStrucTiny~\cite{sibue2026exstructiny} & & & & & \cmark & \cmark & \\
VAREX~\cite{varex2026} & & \cmark & & & & & \\
UniKIE-Bench~\cite{ji2026unikie} & & \cmark & & & & & \\
LongArray-Extract~\cite{extend2026longarray} & \cmark & & \cmark & & & & \\
LongExtractionBench~\cite{micro1longextract} & \cmark & & & & & & \\
ExtractBench~\cite{zhang2026extractbench} & \cmark & \cmark & \cmark & & & \cmark & \cmark \\
\midrule
\textbf{ExtractBench Pro (ours)} & \cmark & \cmark & \cmark & \cmark & \cmark & \cmark & \cmark \\
\bottomrule
\end{tabular*}
\caption{Coverage of extraction challenges and evaluation measures. Checkmarks indicate reported coverage; blank cells indicate no reported coverage. Grounding scores source locations, and cost measures extraction spending. Pro combines these measures with tasks requiring computed values and chart evidence, including printed chart content.}
\label{tab:benchmark-comparison}
\end{table*}

%% file: sections/03_extractbench_pro.tex
\input{sections/construction_options}

\section{Controllable Synthesis of Extraction Tasks}
\label{sec:generation}
To evaluate targeted extraction challenges, we need tasks that reflect real documents and whose content and presentation can be varied to control difficulty. We introduce a synthesis framework that turns structures and conventions from real documents into configurable generation programs (\Cref{fig:construction}). We first formalize the extraction task (\Cref{sec:task-def}). The framework uses a hierarchy of reusable elements, pages, and documents to control content and presentation, while jointly generating documents, extraction schemas, and annotations from shared data (\Cref{sec:generator-functionality}). Coding agents implement and refine these programs using real references and validation feedback (\Cref{sec:agentic-implementation}).

\subsection{Task Definition}
\label{sec:task-def}

Given a PDF document and a user-supplied JSON schema, an extractor system must return all requested values and records that are evident from the document, in the form of a JSON object. The schema specifies the fields, types, and descriptions; the document supplies the evidence needed to determine their values. Descriptions may include clarifying information such as instructions for formatting, when to extract null, or even restrictions on what counts as evidence. Each scalar value (null, text, number, boolean) in the extracted object may also be annotated with one or more page indices or bounding boxes indicating where the evidence for that field is located in the document.

\subsection{From Real Documents to Synthetic Tasks}
\label{sec:generator-functionality}

\paragraph{Grounding in real documents.}
Real reference documents guide both the structures we reproduce and the information we ask an extractor to recover. We encode their layouts and interpretation conventions in reusable, deterministic programs called \emph{generators}. Each generator produces a family of tasks with related schemas.

When executed, a generator begins by filling a fictional underlying data model according to a generator-specific schema. This data model supplies both the values presented in the document and the ground-truth extractions.

\paragraph{Hierarchical composition and control.}
The document is populated from the data model and assembled following a multi-level hierarchy. This separation allows content, visual style, and evidence placement to vary within reusable structures:

\begin{itemize}[leftmargin=1.2em,itemsep=2pt]
\item \textbf{Elements.} Reusable document components such as tables, charts, diagrams, and prose are instantiated with data from the data model and styling from the configuration. We render components using SVG and HTML/CSS. These components are shared across multiple generators.
\item \textbf{Pages and sections.} Layouts arrange elements into pages and sections, controlling where related evidence appears. Additional flavour text and brand-identifiable elements inspired by reference documents are fictionalized and regenerated by the generator to protect the anonymity of the original source. Pagination can cause tables and prose to be broken across pages, which can add extraction difficulty.
\item \textbf{Document.} At the top level, the pages are assembled, potentially with perception-related transformations, and rendered as PDFs.
\end{itemize}

\paragraph{Joint generation from shared data.}
Using seeded randomness, the generator selects names and categories from predetermined fixed pools, and samples dates and numeric values within configured ranges and domain constraints. The document, extraction schema, and annotations are generated jointly from this data model. Derived values are traced in deterministic code, making their derivation reproducible even when extraction requires multiple reasoning steps. Different seeds and configurations produce instances with varied records and presentation.

\input{tables/composition}

\subsection{Generator Development with Coding Agents}
\label{sec:agentic-implementation}

Constructing a generator requires encoding a document's layout, content, and interpretation rules in reusable code. Coding agents make this practical across diverse document families by adapting the generation program to the details of each reference. They inspect real documents, implement reusable components and layouts, execute the generator, and refine its code by comparing rendered outputs with the references. This process combines attention to document-specific detail with explicit controls over content and presentation.

We explicitly supervise generator development through task requirements and validation feedback. Coding agents depend on these external signals to determine what to reproduce and which outputs are acceptable. First, we provide the agent with reference documents and a description of the intended document, extraction schema, or reasoning required for extraction. Within the joint-generation structure and hierarchy described in \Cref{sec:generator-functionality}, the agent develops the data model, rendering components, and answer logic needed for that document family.

Second, we specify validation criteria for the intended use case and use an ensemble of verifiers to assess generated outputs. We combine programmatic checks with agent and human review to assess visual fidelity, schema validity, and whether the rendered document supports the ground-truth extractions and evidence annotations. We return verification failures as feedback, which the agent uses to refine the generator's visual fidelity and task behavior. The resulting program encodes the rendering and answer-computation logic explicitly. We then execute it to generate further instances deterministically while varying their records and presentation. \Cref{sec:pro} describes the requirements and selection criteria used to construct ExtractBench Pro.

\section{ExtractBench Pro}
\label{sec:pro}

ExtractBench Pro contains \prodocs{} documents spanning seven domains and six challenge groups, with a total of \statnum{\propages} pages and \statnum{\provalues} expected scalar values.

\subsection{Benchmark Construction}
\label{sec:composition}

\paragraph{Design goals.}
We construct ExtractBench Pro around three goals: realism, diversity, and difficulty. Documents should reflect real extraction use cases, and schemas should express realistic information needs. Tasks should vary across domains, layouts, and reasoning requirements, while remaining difficult for frontier systems and answerable from the document and schema. Suitable difficult documents are scarce, access can be restricted by privacy and intellectual-property constraints, and manual annotation can introduce errors. Synthesis allows us to control task difficulty and generate annotations from the underlying data.

\input{sections/construction_loop}

\paragraph{Challenge mining.}
We direct coding agents to conduct an automated search through real documents, published definitions, and schemas for structures and interpretation conventions that pose extraction challenges. These include document-specific rules, dependencies across sections, and visual structures that require precise reading. For terminology tasks, we also probe frontier models on candidate rule sets and retain the sets that produce the most errors. We adapt the selected rules for aggregation tasks.

\paragraph{Task construction.}
Using the framework in \Cref{sec:generation}, we prompt coding agents in parallel to implement generators across domains and document types. The generators share a bank of configurable components derived from the mined references and jointly produce documents, schemas, and annotations. We compose challenges so that extracting a value can require several reasoning or perceptual steps. For example, computing a total can require tracing relationships in a diagram and applying their allocation percentages to amounts in a distant table. We vary challenge density and placement, including concentrations beyond those in typical documents. To maintain diversity, we vary layouts and capture conditions and reject documents that are too similar to others.

\paragraph{Challenge calibration.}
We run an ensemble of pilot extractors, including coding agents and specialized APIs, on each generated document and schema, and score their outputs using the metrics in \Cref{sec:metrics}. We target tasks that few or none of these systems extract perfectly and on which most systems fail to resolve the intended challenge. When tasks are too easy, we increase challenge density and repeat the evaluation. We also inspect field-level errors shared by many systems, since these can indicate ambiguous instructions, incorrect annotations, or rendering failures. These findings guide revisions to the generators and tasks.

\paragraph{Dataset composition.}
The resulting benchmark contains \prodocs{} PDF documents produced by \propipelines{} generators, with \statnum{\propages} pages and \statnum{\provalues} expected scalar values. It spans \prondomains{} domains: finance, energy, government, supply chain, healthcare, law, and real estate. Each document is paired with a JSON extraction schema and ground-truth output. \Cref{tab:composition} summarizes the six challenge groups and their document, page, and field counts. Grounding annotations are available for \proboxeddocs{} documents, with bounding boxes covering \proboxedshare\% of expected values across the benchmark.

\subsection{Evaluation Metrics}
\label{sec:metrics}

\paragraph{Value F1.}

We reuse the ExtractBench scorer~\cite{zhang2026extractbench}, which aligns repeated records, normalizes values, and compares scalar fields with the expected output. A value is correct if it matches the expected value or an accepted alternate after normalization. For a document $d$, let $C_d$ be the number of correct scalars, $M_d$ the number of predicted scalar fields, and $G_d$ the number of expected scalar fields in the ground truth. Then Value F1 is

\begin{equation}
F_{1,d}=\frac{2C_d}{M_d+G_d}.
\end{equation}

We report the mean across documents. Failed or empty runs score zero, as do late runs under the main scoring cutoff (\Cref{sec:setup}).

\paragraph{Perfect Value F1.}

Perfect Value F1 reports the proportion of 100 ExtractBench Pro documents with no scored value errors (that is, $F_{1,d} = 1$). Intuitively, it represents the proportion of documents that the user of an extractor system may expect to need no corrections. An error in any part of a document prevents a perfect score, even when many other fields are correct.

\paragraph{Grounding F1.}

Grounding F1 is computed similarly to Value F1, but a scalar counts as correct only if it satisfies the Value F1 rules and is correctly grounded according to the ExtractBench scorer~\cite{zhang2026extractbench}.

\paragraph{Cost per page.}

Cost per page is measured using a document's recorded total cost in USD (including spending on failed and late runs), divided by its page count. Across the entire dataset, we follow ExtractBench conventions and macro-average this across all documents with equal weighting per document.

\FloatBarrier

%% file: sections/construction_options.tex
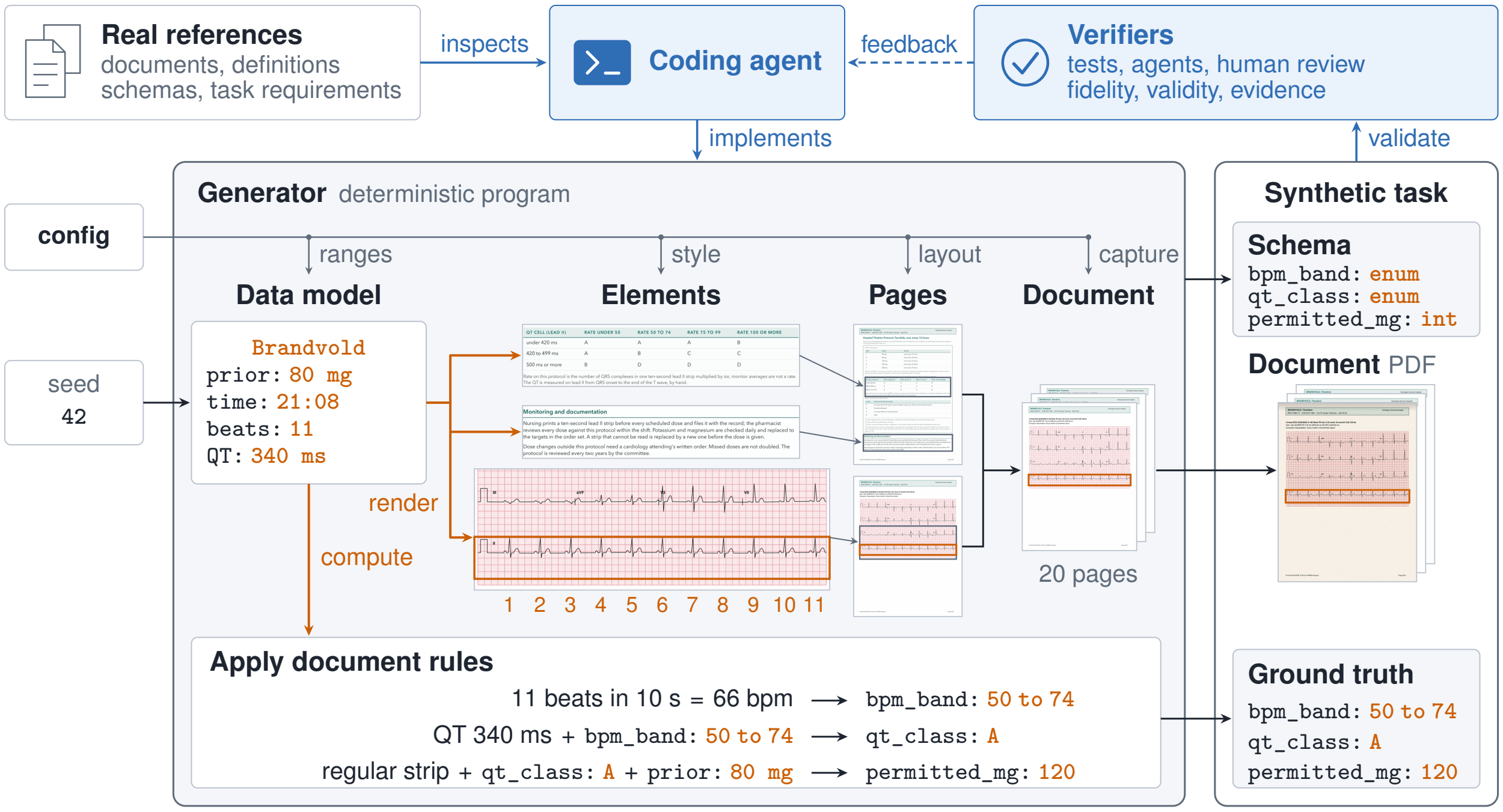
\begin{figure*}[!t]
\centering
\resizebox{\textwidth}{!}{\input{figures/method_overview}}
\caption{Coding agents build configurable generators from real references and revise them using verifier feedback. Each generator produces a document, extraction schema, and ground truth from shared data, with controls over content and presentation. In the highlighted example, the generator draws 11 beats on an ECG strip and computes the expected rate band using the counting rule printed in the document.}
\label{fig:construction}
\label{fig:construction-hierarchy}
\label{fig:construction-paths}
\end{figure*}

%% file: figures/method_overview.tex
\begingroup%
\input{figures/diagram_style}%
\renewcommand{\mvF}{\fontfamily{phv}\fontseries{m}\fontsize{8}{9}\selectfont}%
\renewcommand{\mvT}{\fontfamily{lmtt}\fontseries{m}\fontsize{8}{9}\selectfont\vphantom{Ag}}%
\renewcommand{\mvTB}{\fontfamily{lmtt}\fontseries{b}\fontsize{8}{9}\selectfont\vphantom{Ag}}%
\begin{tikzpicture}[mv picture]
\useasboundingbox (0,0) rectangle (496,266);

\draw[card] (0,0) rectangle (138,38);
\draw[fill=white,draw=mvMuted,line width=0.5pt] (13,6.5) rectangle (25,22.5);
\draw[fill=white,draw=mvMuted,line width=0.5pt,line join=round]
  (7,11.5) -- (16,11.5) -- (20,15.5) -- (20,30.5) -- (7,30.5) -- cycle;
\draw[mvMuted,line width=0.5pt,line join=round] (16,11.5) -- (16,15.5) -- (20,15.5);
\foreach \y/\x in {19.5/17.5,23/17.5,26.5/14.5}
  \draw[mvMuted,line width=0.5pt] (9.5,\y) -- (\x,\y);
\node[anchor=west] at (32,10.5) {\mvB{9}Real references};
\node[anchor=west,text=mvMuted] at (32,20.3) {documents, definitions};
\node[anchor=west,text=mvMuted] at (32,28.8) {schemas, task requirements};

\draw[card,draw=mvBlue,fill=mvBlueFill] (181,0) rectangle (279,38);
\fill[mvBlue,rounded corners=2pt] (189,11.5) rectangle (208,26.5);
\draw[white,line width=1pt,line cap=round,line join=round] (193.5,15.5) -- (197,19) -- (193.5,22.5);
\draw[white,line width=1pt,line cap=round] (199.5,22.5) -- (204,22.5);
\node[anchor=west,text=mvBlue] at (214,19) {\mvB{9}Coding agent};

\draw[card,draw=mvBlue,fill=mvBlueFill] (322,0) rectangle (496,38);
\draw[mvBlue,line width=0.9pt] (339,19) circle[radius=7.5pt];
\draw[mvBlue,line width=1.1pt,line cap=round,line join=round] (335.2,19.4) -- (338,22.4) -- (343,16);
\node[anchor=west,text=mvBlue] at (353,10.5) {\mvB{9}Verifiers};
\node[anchor=west,text=mvBlue] at (353,20.3) {tests, agents, human review};
\node[anchor=west,text=mvBlue] at (353,28.8) {fidelity, validity, evidence};

\draw[dev] (138,19) -- (180,19);
\node[text=mvBlue] at (159.5,13.5) {inspects};
\draw[loop] (322,19) -- (280,19);
\node[text=mvBlue] at (300.5,13.5) {feedback};
\draw[dev] (230,38) -- (230,51.5);
\node[anchor=west,text=mvBlue] at (234,44.8) {implements};
\draw[dev] (449,52) -- (449,38.5);
\node[anchor=west,text=mvBlue] at (453,44.8) {validate};

\draw[draw=mvMuted,line width=0.6pt,rounded corners=4pt,fill=mvPanel] (56,52) rectangle (392,266);
\node[anchor=west] at (64,63) {\mvB{9}Generator\hspace{4pt}{\mvS{8}\color{mvMuted}deterministic program}};

\draw[card] (0,66) rectangle (46,88);
\node at (23,77) {\mvB{8}config};
\draw[card] (0,118) rectangle (46,146);
\node[text=mvMuted] at (23,126.5) {seed};
\node at (23,137) {\fontfamily{lmtt}\fontseries{b}\fontsize{8}{8}\selectfont\vphantom{Ag}42};
\draw[knob] (46,77) -- (360,77);
\foreach \x/\label in {101/ranges,218/style,300/layout,360/capture} {
  \draw[knob,small tip] (\x,77) -- (\x,89.5);
  \fill[mvMuted] (\x,77) circle[radius=0.9pt];
  \node[anchor=west,text=mvMuted] at (\x+3.5,83.3) {\label};
}
\foreach \x/\label in {101/Data model,218/Elements,300/Pages,360/Document}
  \node at (\x,97) {\mvB{9}\label};

\draw[card] (62,105) rectangle (140,159);
\node at (101,114) {\mvTB\color{mvOrange}Brandvold};
\node[anchor=west] at (67,123) {\mvT prior:\hspace{2pt}{\mvTB\color{mvOrange}80 mg}};
\node[anchor=west] at (67,132) {\mvT time:\hspace{2pt}{\mvTB\color{mvOrange}21:08}};
\node[anchor=west] at (67,141) {\mvT beats:\hspace{2pt}{\mvTB\color{mvOrange}11}};
\node[anchor=west] at (67,150) {\mvT QT:\hspace{2pt}{\mvTB\color{mvOrange}340 ms}};
\draw[flow] (46,132) -- (61.5,132);

\draw[draw=mvOrange,line width=0.8pt] (140,132) -- (148,132);
\draw[draw=mvOrange,line width=0.8pt,line cap=round] (148,116.3) -- (148,176.8);
\foreach \y in {116.3,141.7} \draw[data] (148,\y) -- (171.5,\y);
\draw[data] (148,176.8) -- (155.5,176.8);
\node[anchor=east,text=mvOrange] at (144,166) {render};
\node[anchor=north west] at (172,106) {\includegraphics[width=92pt]{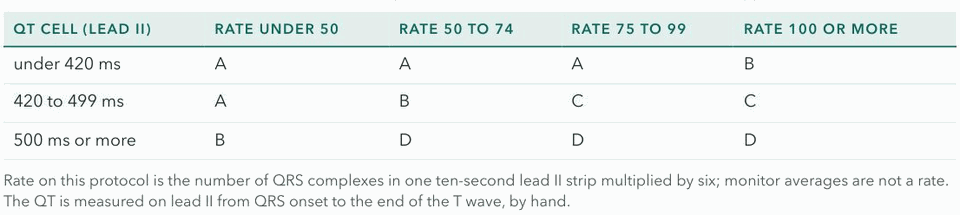}};
\draw[mvLine,line width=0.35pt] (172,106) rectangle (264,126.6);
\node[anchor=north west] at (172,133) {\includegraphics[width=92pt]{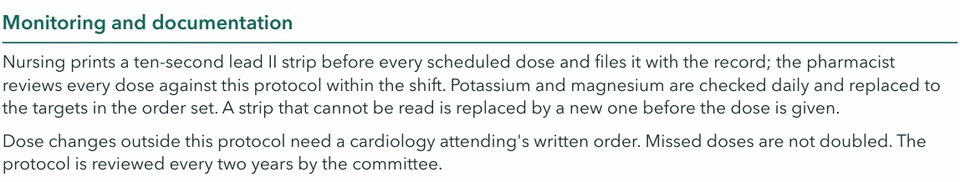}};
\draw[mvLine,line width=0.35pt] (172,133) rectangle (264,150.5);
\node[anchor=north west] at (156,154) {\includegraphics[width=118pt]{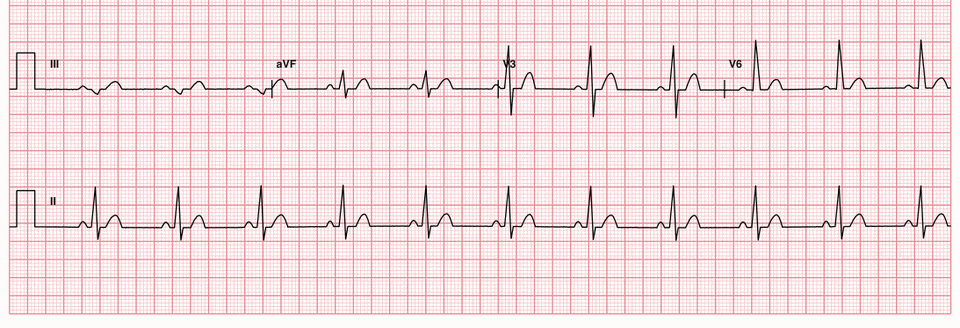}};
\draw[mvLine,line width=0.35pt] (156,154) rectangle (274,194.27);
\draw[mvOrange,line width=0.7pt] (156,176.57) rectangle (274,190.55);
\foreach \x [count=\beat] in {167.54,177.80,187.93,198.07,208.33,218.46,228.46,238.59,248.73,258.99,268.99}
  \node[text=mvOrange] at (\x,199.5) {\mvS{7}\beat};

\node[anchor=north west] at (282,106) {\includegraphics[width=36pt]{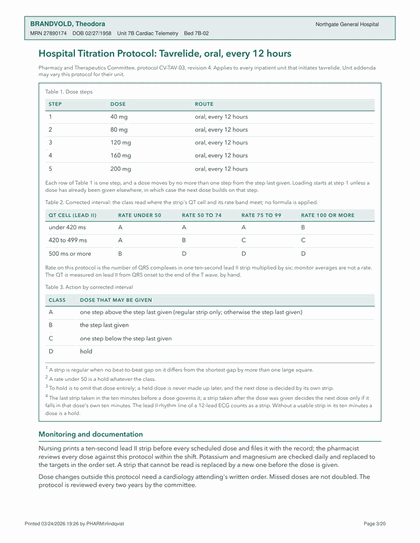}};
\draw[mvLine,line width=0.35pt] (282,106) rectangle ++(36,{36*1.29375});
\node[anchor=north west] at (282,156.5) {\includegraphics[width=36pt]{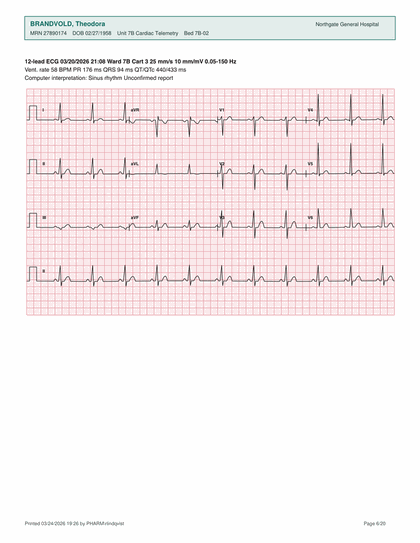}};
\draw[mvLine,line width=0.35pt] (282,156.5) rectangle ++(36,{36*1.29375});
\draw[mvMuted,line width=0.45pt] (285.8,123.6) rectangle (314.2,130);
\draw[mvMuted,line width=0.45pt] (285.2,142.6) rectangle (314.8,148.1);
\draw[mvMuted,line width=0.45pt] (284,173) rectangle (316.1,184);
\draw[mvOrange,line width=0.6pt] (284,179.2) rectangle (316.1,182.6);
\draw[place] (264,116.3) -- (285.3,126.6);
\draw[place] (264,141.7) -- (284.7,145.2);
\draw[place] (274,176.8) -- (283.5,178.3);

\draw[draw=mvInk,line width=0.6pt] (318,129.3) -- (325,129.3) -- (325,179.8) -- (318,179.8);
\draw[flow] (325,154.5) -- (337.5,154.5);
\node[anchor=north west] at (344,126) {\includegraphics[width=38pt]{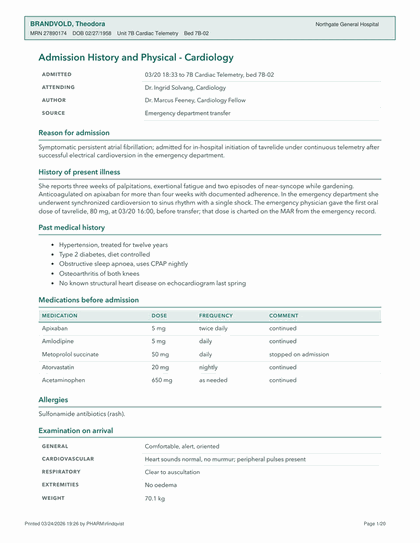}};
\draw[mvLine,line width=0.35pt] (344,126) rectangle ++(38,{38*1.29375});
\node[anchor=north west] at (341,129) {\includegraphics[width=38pt]{method/page-3.png}};
\draw[mvLine,line width=0.35pt] (341,129) rectangle ++(38,{38*1.29375});
\node[anchor=north west] at (338,132) {\includegraphics[width=38pt]{method/page-6-clean.png}};
\draw[mvLine,line width=0.35pt] (338,132) rectangle ++(38,{38*1.29375});
\draw[mvOrange,line width=0.6pt] (340.1,155.9) rectangle (374,159.5);
\node[text=mvMuted] at (360,189.5) {20 pages};

\draw[data] (101,159) -- (101,209.5);
\node[anchor=west,text=mvOrange] at (105,184) {compute};
\draw[card] (62,210) rectangle (384,260);
\node[anchor=west] at (68,219) {\mvB{9}Apply document rules};
\node[anchor=east] at (262,231) {11 beats in 10 s\hspace{3pt}=\hspace{3pt}66 bpm};
\node[anchor=east] at (262,243) {QT 340 ms\hspace{3pt}+\hspace{3pt}{\mvT bpm\_band:\hspace{2pt}{\mvTB\color{mvOrange}50\hspace{2pt}to\hspace{2pt}74}}};
\node[anchor=east] at (262,255) {regular strip\hspace{3pt}+\hspace{3pt}{\mvT qt\_class:\hspace{2pt}{\mvTB\color{mvOrange}A}}\hspace{3pt}+\hspace{3pt}{\mvT prior:\hspace{2pt}{\mvTB\color{mvOrange}80 mg}}};
\foreach \y in {231,243,255}
  \draw[flow] (268,\y) -- (280,\y);
\node[anchor=west] at (286,231) {\mvT bpm\_band:\hspace{2pt}{\mvTB\color{mvOrange}50\hspace{2pt}to\hspace{2pt}74}};
\node[anchor=west] at (286,243) {\mvT qt\_class:\hspace{2pt}{\mvTB\color{mvOrange}A}};
\node[anchor=west] at (286,255) {\mvT permitted\_mg:\hspace{2pt}{\mvTB\color{mvOrange}120}};

\draw[draw=mvMuted,line width=0.6pt,rounded corners=4pt,fill=white] (402,52) rectangle (496,266);
\node at (449,63) {\mvB{9}Synthetic task};
\draw[card,fill=mvPanel] (408,72) rectangle (490,110);
\node[anchor=west] at (413,80.5) {\mvB{9}Schema};
\node[anchor=west] at (413,89.5) {\mvT bpm\_band:\hspace{2pt}{\mvTB\color{mvOrange}enum}};
\node[anchor=west] at (413,97) {\mvT qt\_class:\hspace{2pt}{\mvTB\color{mvOrange}enum}};
\node[anchor=west] at (413,104.5) {\mvT permitted\_mg:\hspace{2pt}{\mvTB\color{mvOrange}int}};
\draw[flow] (392,91) -- (407.5,91);

\node[anchor=west] at (413,120) {\mvB{9}Document\hspace{2.5pt}{\mvS{8}\color{mvMuted}PDF}};
\node[anchor=north west] at (429,126) {\includegraphics[width=46pt]{method/page-1.png}};
\draw[mvLine,line width=0.35pt] (429,126) rectangle ++(46,{46*1.29375});
\node[anchor=north west] at (426,129) {\includegraphics[width=46pt]{method/page-3.png}};
\draw[mvLine,line width=0.35pt] (426,129) rectangle ++(46,{46*1.29375});
\node[anchor=north west] at (423,132) {\includegraphics[width=46pt]{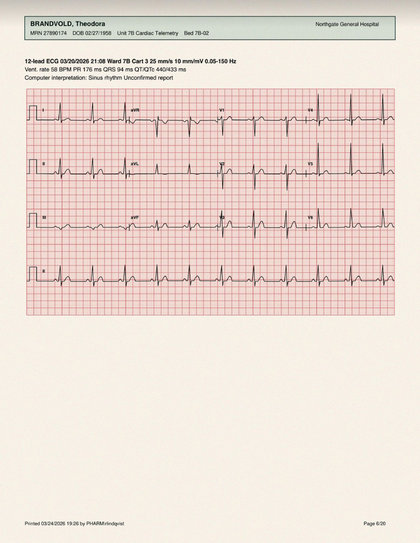}};
\draw[mvLine,line width=0.35pt] (423,132) rectangle ++(46,{46*1.29375});
\draw[mvOrange,line width=0.6pt] (425.5,161) rectangle (466.6,165.2);
\draw[flow] (382.5,154.5) -- (422.5,154.5);

\draw[card,fill=mvPanel] (408,214) rectangle (490,260);
\node[anchor=west] at (413,223) {\mvB{9}Ground truth};
\node[anchor=west] at (413,235) {\mvT bpm\_band:\hspace{2pt}{\mvTB\color{mvOrange}50\hspace{2pt}to\hspace{2pt}74}};
\node[anchor=west] at (413,245) {\mvT qt\_class:\hspace{2pt}{\mvTB\color{mvOrange}A}};
\node[anchor=west] at (413,255) {\mvT permitted\_mg:\hspace{2pt}{\mvTB\color{mvOrange}120}};
\draw[flow] (384,237) -- (407.5,237);
\end{tikzpicture}
\endgroup

%% file: tables/composition.tex
\begin{table*}[t]
\centering
\small
\setlength{\tabcolsep}{3pt}
\begin{tabular*}{\textwidth}{@{\extracolsep{\fill}}llrrrr@{}}
\toprule
\textbf{Challenge group} & \textbf{Main elements} & \textbf{Generators} & \textbf{Docs} & \textbf{Pages} & \textbf{Values} \\
\midrule
Interpretation rules & \textbf{Rules}, aggregation & 18 & 22 & \numprint{875} & \numprint{121287} \\
Custom terminology & \textbf{Jargon}, rules & 14 & 14 & \numprint{606} & \numprint{15242} \\
Conditional aggregation & \textbf{Aggregation}, rules & 12 & 12 & \numprint{225} & \numprint{409} \\
Relational diagrams & \textbf{Relational diagrams}, rules, aggregation & 17 & 20 & \numprint{352} & \numprint{2349} \\
Charts and maps & \textbf{Charts and maps}, rules & 25 & 25 & \numprint{557} & \numprint{1347} \\
Long single pages & \textbf{Extreme aspect ratios}, charts and maps & 4 & 7 & \numprint{7} & \numprint{3507} \\
\midrule
\textbf{All} & & 90 & 100 & \numprint{2622} & \numprint{144141} \\
\bottomrule
\end{tabular*}
\caption{Dataset composition across challenge groups. Values are scalar fields in the expected output, including nulls.}
\label{tab:composition}
\end{table*}

%% file: sections/construction_loop.tex
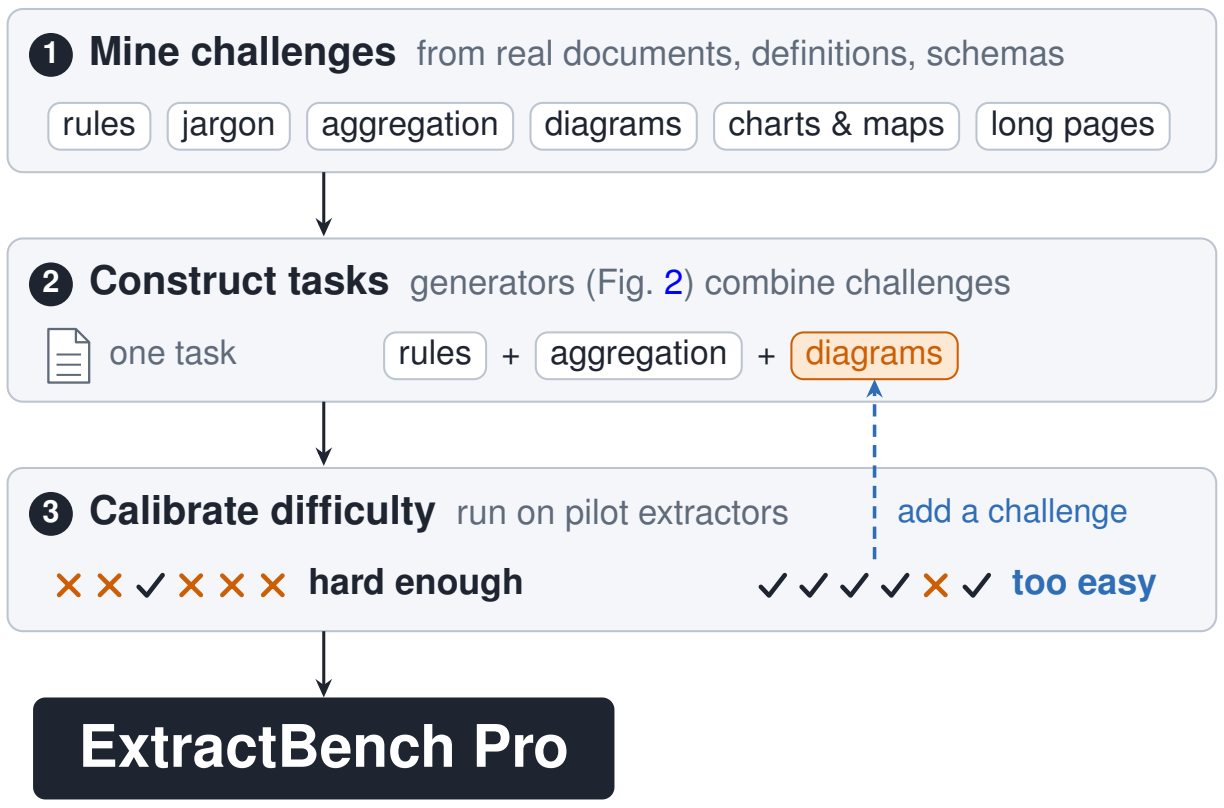
\begin{figure}[t]
\centering
\resizebox{\columnwidth}{!}{\input{figures/construction_loop}}
\caption{The construction process for the ExtractBench Pro benchmark, starting with challenge mining followed by a feedback cycle to construct appropriately-difficult extraction tasks.}
\label{fig:calibration}
\end{figure}

%% file: figures/construction_loop.tex
\begingroup%
\input{figures/diagram_style}%
\newcommand{\mvmarks}[2]{%
  \foreach \x/\ok in {#2} {%
    \ifnum\ok=1 \draw[mvInk,line width=0.9pt,line cap=round,line join=round]
      (\x-2.3,#1+0.3) -- (\x-0.7,#1+2) -- (\x+2.4,#1-2);%
    \else \draw[mvOrange,line width=0.9pt,line cap=round]
      (\x-1.9,#1-1.9) -- (\x+1.9,#1+1.9) (\x-1.9,#1+1.9) -- (\x+1.9,#1-1.9);\fi}}%
\newcommand{\mvdoc}[2]{%
  \draw[fill=white,draw=mvMuted,line width=0.45pt,line join=round]
    (#1,#2) -- ++(5.4,0) -- ++(2.6,2.6) -- ++(0,7.9) -- ++(-8,0) -- cycle;
  \draw[mvMuted,line width=0.45pt,line join=round] (#1+5.4,#2) -- ++(0,2.6) -- ++(2.6,0);
  \foreach \dy in {5,7,9} \draw[mvMuted,line width=0.4pt] (#1+1.6,#2+\dy) -- (#1+6.4,#2+\dy);}%
\newcommand{\mvstep}[4]{%
  \fill[mvInk] (8.5,#1) circle[radius=4.3pt];
  \node[text=white] at (8.5,#1) {\mvB{6}#2};
  \node[anchor=west] at (16,#1) {\mvB{7.8}#3\hspace{4pt}{\mvS{6.8}\color{mvMuted}#4}};}%
\begin{tikzpicture}[mv picture,
  stage/.style={draw=mvLine,line width=0.4pt,rounded corners=3pt,fill=mvPanel},
  chip/.style={draw=mvLine,line width=0.4pt,rounded corners=2.4pt,fill=white,
    inner xsep=2.8pt,inner ysep=1.5pt,font=\fontfamily{phv}\fontseries{m}\fontsize{6.6}{6.6}\selectfont},
  added/.style={chip,draw=mvOrange,fill=mvOrangeFill,text=mvOrange}]
\useasboundingbox (0,0) rectangle (237,158);

\draw[stage] (0,0) rectangle (237,32);
\mvstep{9}{1}{Mine challenges}{from real documents, definitions, schemas}
\node[chip,anchor=west] (m1) at (8,23) {rules};
\node[chip,right=3.4pt of m1] (m2) {jargon};
\node[chip,right=3.4pt of m2] (m3) {aggregation};
\node[chip,right=3.4pt of m3] (m4) {diagrams};
\node[chip,right=3.4pt of m4] (m5) {charts \& maps};
\node[chip,right=3.4pt of m5] (m6) {long pages};
\draw[flow] (62,32) -- (62,44.6);

\draw[stage] (0,45) rectangle (237,77);
\mvstep{54}{2}{Construct tasks}{generators (Fig.~\ref{fig:construction}) combine challenges}
\mvdoc{8}{62.75}
\node[anchor=west,text=mvMuted] at (20,68) {\mvS{6.6}one task};
\node[added] (b3) at (170,68) {diagrams};
\node[left=3pt of b3] (p2) {\mvS{6.4}+};
\node[chip,left=3pt of p2] (b2) {aggregation};
\node[left=3pt of b2] (p1) {\mvS{6.4}+};
\node[chip,left=3pt of p1] (b1) {rules};
\draw[flow] (62,77) -- (62,89.6);

\draw[stage] (0,90) rectangle (237,122);
\mvstep{99}{3}{Calibrate difficulty}{run on pilot extractors}
\mvmarks{113}{12/0,20/0,28/1,36/0,44/0,52/0}
\node[anchor=west] at (59,113) {\mvB{7}hard enough};
\mvmarks{113}{150/1,158/1,166/1,174/1,182/0,190/1}
\node[anchor=west,text=mvBlue] at (197,113) {\mvB{7}too easy};

\draw[loop] (170,107.9) -- (b3.south);
\node[anchor=west,text=mvBlue] at (174.5,99) {add a challenge};

\node[anchor=north,fill=mvInk,rounded corners=2.4pt,minimum width=114pt,minimum height=20pt,inner sep=0pt] (pro)
  at (62,135) {};
\node[text=white,inner sep=0pt,yshift=-0.8pt] at (pro.center) {\mvB{11.5}ExtractBench Pro};
\draw[flow] (62,122) -- (pro.north);
\end{tikzpicture}
\endgroup

%% file: sections/04_experiments.tex
\section{Experiments}
\label{sec:results}

\begin{figure*}[!t]
\centering
\includegraphics[width=\textwidth]{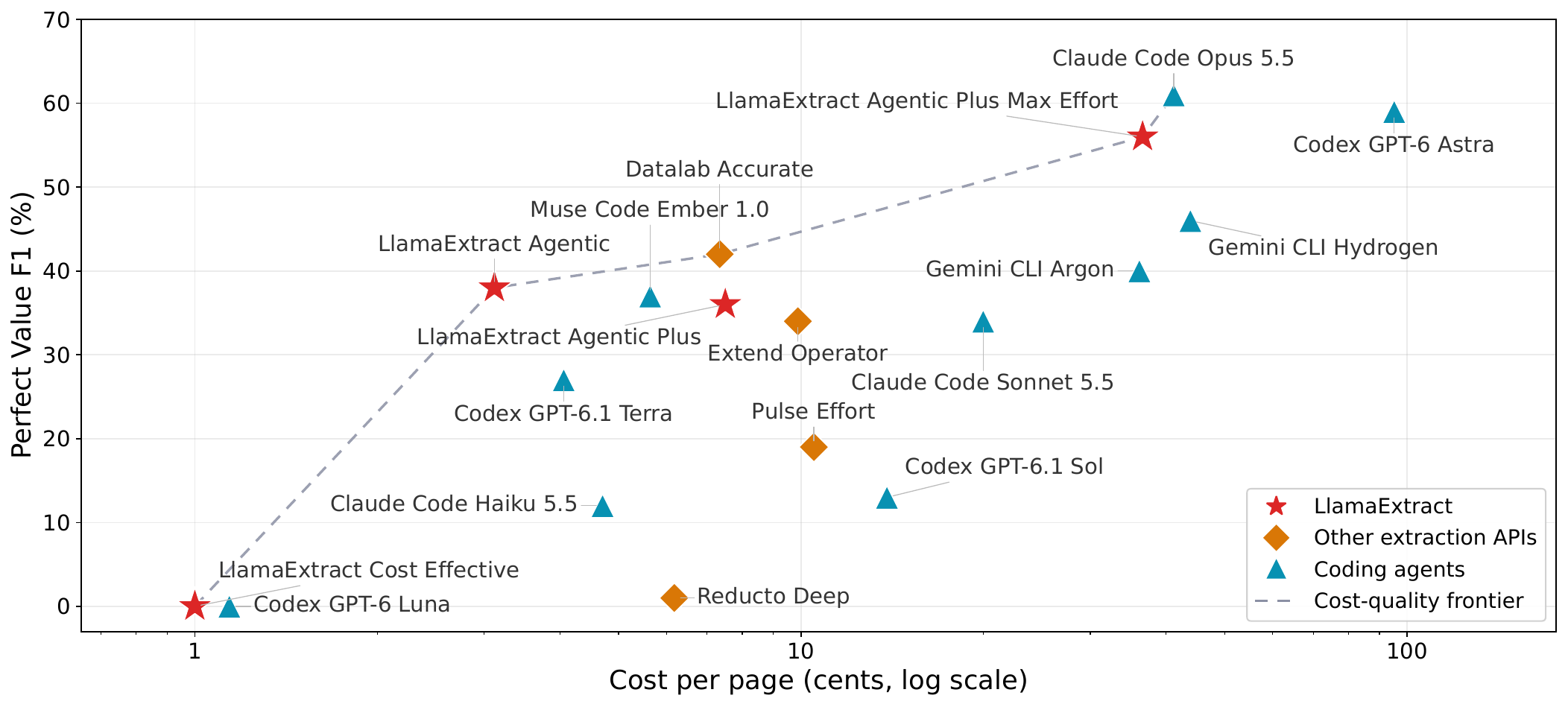}
\caption{Perfect Value F1 and mean document-level cost per page. The dashed line joins the observed cost-quality frontier.}
\label{fig:quality-cost}
\end{figure*}

\subsection{Setup}
\label{sec:setup}

We evaluate ten coding-agent configurations from OpenAI, Anthropic, Google, and Meta, alongside eight extraction API configurations across five commercial providers: four tiers of LlamaExtract (Cost Effective, Agentic, Agentic Plus, and Agentic Plus Max Effort) and the flagship extraction tier from each of Datalab, Extend, Pulse, and Reducto (\Cref{tab:results-overall}). Coding agents receive the PDF and schema as files and are asked to return evidence. Each agent runs in an isolated environment, and every step is recorded for later analysis. Meanwhile, APIs receive the same inputs through their extraction interfaces. All systems are evaluated with a three-hour limit per document, including retries. Failed runs, empty outputs, and runs exceeding this limit remain in the accuracy denominator and score zero.

\input{tables/results_overall}
\input{tables/results_by_challenge}

\subsection{Overall Results}
\label{sec:overall}
\label{sec:cost}
\label{sec:grounding}

Overall, the top extraction systems on ExtractBench Pro extract at most \bestPerfect{}\% of documents perfectly: \opPerfect{} by Claude Opus 5.5, \axPerfect{} by Codex GPT-6 Astra, and \amPerfect{} by LlamaExtract Agentic Plus Max. Interestingly, it is not the same documents which are extracted perfectly by all systems, demonstrating clear advantages and tradeoffs between systems; combined, \topThreeUnionPerfect{} documents are extracted perfectly by at least one of the 3 aforementioned systems, and \proUnionPerfect{} by at least one of the \pronsystems{} systems evaluated in this paper. Therefore, there is room for future advancements combining those strengths to surpass the \bestPerfect{}\% frontier.

For comparison, applying ExtractBench's 30-minute scoring cutoff to the same saved outputs reduces Opus's Perfect Value F1 from \opPerfect\% to \opPerfectThirtyMinute\%, and Value F1 from \opValue\% to \opValueThirtyMinute\% (\Cref{tab:results-overall-no-limit}). All extraction APIs except Datalab retain their scores under this cutoff, consistent with their generally lower latency, while non-Codex coding agents incur the largest accuracy losses.

High Value F1 can coexist with low Perfect Value F1. On weaker-performing systems, perfect document rate tends to decline faster than average Value F1, demonstrating that it is far more difficult to extract a document perfectly (thereby solving all challenges present in that document) than it is to extract most of the values correctly. For example, although Pulse Effort achieves about 75\% of the Value F1 of the leading extractors, it extracts less than one-third as many documents perfectly.

Viewing results against cost clarifies the tradeoff between coding agents and extraction APIs (\Cref{fig:quality-cost}). Coding agents reach the highest accuracy by writing custom scripts without a per-page budget, whereas specialized extraction APIs trade some ceiling for lower, predictable per-page pricing. \costFrontierNames{} form the observed cost-quality frontier. LlamaExtract Agentic outperforms most competing APIs and mid-tier coding agents, while Agentic Plus Max achieves \amPerfect{}\% Perfect Value F1 and \amValue{}\% Value F1 at less than 40\% of Astra's per-page cost.

\subsection{Results by Challenge}
\label{sec:by-challenge}
\label{sec:failures}

Separating results by challenge group explains why no single system dominates (\Cref{tab:results-by-challenge}). By Perfect Value F1, Claude Opus 5.5 leads relational diagrams and charts and maps, GPT-6 Astra leads long single pages and ties Datalab Accurate on conditional aggregation, and LlamaExtract Agentic Plus Max leads custom terminology. Agentic Plus Max, Claude Opus, and Gemini Hydrogen tie on interpretation rules. Within LlamaExtract, the lower-cost Agentic and Agentic Plus tiers already perform strongly on text- and rule-heavy splits, while Max Effort makes the biggest difference on diagrams, charts, and long sheets. No single system leads every challenge group.

\subsection{Trace Analysis}
\label{sec:trace-analysis}

We inspected saved tool calls, generated scripts, and intermediate outputs by coding agents. The agents constructed sophisticated task-specific programs that combined extracted observations with document-defined rules. For example, on a fund-structure document, Codex Astra transcribed ownership relationships from handwritten diagrams, extracted funding amounts from tables, and propagated contributions through the resulting structure using exact-decimal arithmetic.

Visual extraction also involved explicit measurement and image processing. On a scanned time--location chart, Astra corrected image skew, traced activity lines, and mapped their endpoints to the document's week and distance rulers. It then used geometric intersections to calculate progress and crossing counts.

Despite these impressive programs, coding agents are still susceptible to multi-hop reasoning, as a single mistake made at any stage is enough to render the resulting extractions incorrect. A common mistake that triggered this failure was the incorrect interpretation of a rule, such as inclusion in an aggregation condition.

Coding agents also still struggled to provide evidence for derived answers. Systems are biased towards citing labels that appear in literal form in the document, while omitting calculated fields. A correct grounding of those answers would require identifying the many source values and rules used in the calculation.

\subsection{Realism}
\label{sec:realism}

To evaluate the realism of the generated documents, a small study was conducted asking six non-expert volunteers to compare ExtractBench and ExtractBench Pro to real customer-submitted documents sent to a specialized extraction API. The study presented the user with a pair of pages, consisting of one page sampled from real extraction traffic, versus one page sampled from the dataset. Each participant was asked to identify the real page, or abstain. Detection rates were 23.8\% for ExtractBench and 37.9\% for ExtractBench Pro. Therefore, while ExtractBench Pro documents are easier than ExtractBench to discriminate from real documents, it is still well below 50\% (a random coin toss), demonstrating a reasonable degree of realism.

From qualitative inspection of the pages, the main gaps of fidelity were found to be handwriting and the lack of brand intellectual property such as logos. Synthetic handwriting was easy to distinguish from real handwriting due to the use of fonts. Meanwhile, brand IP is questionable to use in fake documents due to privacy and confidentiality concerns, and was deliberately avoided in this benchmark.

%% file: tables/results_overall.tex
\begin{table}[t]
\centering
\small
\setlength{\tabcolsep}{2pt}
\begin{tabular}{@{}lrrrr@{}}
\toprule
\textbf{System} & \makecell{\textbf{Perfect}\\\textbf{Value F1}} & \makecell{\textbf{Value}\\\textbf{F1}} & \makecell{\textbf{Grounding}\\\textbf{F1}} & \textbf{\textcent/page} \\
\midrule
\multicolumn{5}{@{}l}{\textit{Coding agents}} \\
\hspace{0.7em}Codex GPT-6 Astra & \underline{59.0} & \textbf{96.3} & \textbf{72.2} & 95.2 \\
\hspace{0.7em}Codex GPT-6.1 Sol & 13.0 & 75.8 & 65.0 & 13.9 \\
\hspace{0.7em}Codex GPT-6 Luna & 0.0 & 53.1 & 47.6 & 1.1 \\
\hspace{0.7em}Codex GPT-6.1 Terra & 27.0 & 74.2 & 60.2 & 4.1 \\
\hspace{0.7em}CC Opus 5.5 & \textbf{61.0} & \underline{93.1} & 60.3 & 41.2 \\
\hspace{0.7em}CC Sonnet 5.5 & 34.0 & 86.8 & 47.3 & 20.0 \\
\hspace{0.7em}CC Haiku 5.5 & 12.0 & 60.2 & 42.7 & 4.7 \\
\hspace{0.7em}Muse Code Ember 1.0 & 37.0 & 81.2 & 49.4 & 5.6 \\
\hspace{0.7em}Gemini CLI Argon & 40.0 & 61.8 & 47.0 & 36.2 \\
\hspace{0.7em}Gemini CLI Hydrogen & 46.0 & 80.3 & 59.1 & 43.9 \\
\multicolumn{5}{@{}l}{\textit{Extraction APIs}} \\
\hspace{0.7em}LE Agentic Plus Max & 56.0 & 92.7 & \underline{66.2} & 36.6 \\
\hspace{0.7em}LE Agentic Plus & 36.0 & 81.2 & 62.4 & 7.5 \\
\hspace{0.7em}LE Agentic & 38.0 & 80.6 & 61.4 & 3.1 \\
\hspace{0.7em}LE Cost Effective & 0.0 & 47.2 & 49.4 & 1.0 \\
\hspace{0.7em}Datalab Accurate & 42.0 & 82.5 & 46.5 & 7.3 \\
\hspace{0.7em}Extend Operator & 34.0 & 73.9 & 57.7 & 9.9 \\
\hspace{0.7em}Pulse Effort & 19.0 & 72.1 & 19.2 & 10.5 \\
\hspace{0.7em}Reducto Deep & 1.0 & 58.4 & 45.3 & 6.2 \\
\bottomrule
\end{tabular}
\caption{Overall results. Bold and underline mark the highest and second-highest scores in each quality column. LE: LlamaExtract; CC: Claude Code.}
\label{tab:results-overall}
\end{table}

%% file: tables/results_by_challenge.tex
\begin{table*}[!t]
\centering
\small
\setlength{\tabcolsep}{2.3pt}
\begin{tabular}{@{}l*{6}{rr}@{}}
\toprule
\textbf{System} & \multicolumn{2}{c}{\makecell{\textbf{Interpretation}\\\textbf{rules}}} & \multicolumn{2}{c}{\makecell{\textbf{Custom}\\\textbf{terminology}}} & \multicolumn{2}{c}{\makecell{\textbf{Conditional}\\\textbf{aggregation}}} & \multicolumn{2}{c}{\makecell{\textbf{Relational}\\\textbf{diagrams}}} & \multicolumn{2}{c}{\makecell{\textbf{Charts and}\\\textbf{maps}}} & \multicolumn{2}{c}{\makecell{\textbf{Long single}\\\textbf{pages}}} \\
\cmidrule(lr){2-3}\cmidrule(lr){4-5}\cmidrule(lr){6-7}\cmidrule(lr){8-9}\cmidrule(lr){10-11}\cmidrule(lr){12-13}
 & Perfect & F1 & Perfect & F1 & Perfect & F1 & Perfect & F1 & Perfect & F1 & Perfect & F1 \\
\midrule
\multicolumn{13}{@{}l}{\textit{Coding agents}} \\
\quad Codex GPT-6 Astra & \underline{27.3} & 97.2 & 64.3 & \textbf{99.5} & \textbf{66.7} & \textbf{89.3} & \underline{80.0} & \underline{98.0} & 64.0 & 94.7 & \textbf{57.1} & \textbf{99.7} \\
\quad Codex GPT-6.1 Sol & 9.1 & 91.4 & 14.3 & 86.6 & 16.7 & 46.6 & 25.0 & 75.1 & 4.0 & 65.6 & 14.3 & 93.6 \\
\quad Codex GPT-6 Luna & 0.0 & 71.9 & 0.0 & 73.4 & 0.0 & 30.4 & 0.0 & 39.6 & 0.0 & 55.0 & 0.0 & 23.4 \\
\quad Codex GPT-6.1 Terra & 9.1 & 67.5 & 14.3 & 89.2 & 25.0 & 63.6 & 30.0 & 74.4 & 56.0 & 87.7 & 0.0 & 34.6 \\
\quad CC Opus 5.5 & \textbf{36.4} & 97.5 & 71.4 & 98.5 & 33.3 & 60.9 & \textbf{90.0} & \textbf{98.2} & \textbf{80.0} & \textbf{97.8} & 14.3 & 92.1 \\
\quad CC Sonnet 5.5 & 22.7 & 91.0 & 14.3 & 96.0 & 33.3 & 65.3 & 60.0 & 89.5 & 44.0 & 88.0 & 0.0 & 80.4 \\
\quad CC Haiku 5.5 & 9.1 & 75.1 & 0.0 & 73.5 & 25.0 & 56.1 & 25.0 & 42.0 & 8.0 & 66.3 & 0.0 & 24.1 \\
\quad Muse Code Ember 1.0 & 18.2 & 90.6 & 35.7 & 93.1 & 16.7 & 54.3 & 50.0 & 69.5 & 56.0 & 89.2 & \underline{28.6} & 79.4 \\
\quad Gemini CLI Argon & 9.1 & 31.3 & \underline{78.6} & 85.7 & 16.7 & 48.7 & 55.0 & 67.4 & 56.0 & 85.5 & 0.0 & 32.1 \\
\quad Gemini CLI Hydrogen & \textbf{36.4} & 68.0 & 64.3 & 78.5 & 41.7 & 71.7 & 60.0 & 80.0 & 44.0 & 91.4 & 14.3 & \underline{98.8} \\
\multicolumn{13}{@{}l}{\textit{Extraction APIs}} \\
\quad LE Agentic Plus Max Effort & \textbf{36.4} & \textbf{99.7} & \textbf{85.7} & \textbf{99.5} & 33.3 & 60.1 & 70.0 & 96.0 & \underline{72.0} & \underline{95.7} & 0.0 & 92.8 \\
\quad LE Agentic Plus & 13.6 & 97.1 & 64.3 & 98.4 & 33.3 & 61.2 & 35.0 & 65.9 & 52.0 & 88.7 & 0.0 & 48.4 \\
\quad LE Agentic & 18.2 & 99.0 & \underline{78.6} & \underline{98.7} & \underline{50.0} & 70.4 & 15.0 & 67.7 & 56.0 & 90.1 & 0.0 & 7.1 \\
\quad LE Cost Effective & 0.0 & 75.0 & 0.0 & 80.8 & 0.0 & 33.4 & 0.0 & 29.8 & 0.0 & 37.6 & 0.0 & 0.1 \\
\quad Datalab Accurate & 18.2 & 93.8 & \underline{78.6} & 97.8 & \textbf{66.7} & \underline{86.0} & 20.0 & 69.9 & 60.0 & \underline{95.7} & 0.0 & 0.0 \\
\quad Extend Operator & \underline{27.3} & \underline{99.6} & 64.3 & 97.6 & 16.7 & 59.6 & 0.0 & 41.2 & 68.0 & 91.6 & 0.0 & 0.0 \\
\quad Pulse Effort & 9.1 & 93.8 & 57.1 & 89.3 & 25.0 & 61.0 & 15.0 & 65.0 & 12.0 & 65.0 & 0.0 & 34.1 \\
\quad Reducto Deep & 0.0 & 79.5 & 0.0 & 85.1 & 0.0 & 51.4 & 0.0 & 38.3 & 4.0 & 56.1 & 0.0 & 17.1 \\
\bottomrule
\end{tabular}
\caption{Perfect Value F1 / Value F1 by challenge, in percent. Bold and underline mark the highest and second-highest scores in each column. LE: LlamaExtract; CC: Claude Code.}
\label{tab:results-by-challenge}
\end{table*}

%% file: sections/05_conclusion.tex
\section{Conclusion}
\label{sec:conclusion}

We introduce ExtractBench Pro, a benchmark of 100 documents built using configurable, deterministic document generators. This approach lets us compose extraction challenges that remain difficult for frontier systems while generating documents, schemas, and ground truth together. The strongest system extracts only \bestPerfect{}\% of documents perfectly, while specialized tiers such as LlamaExtract narrow the accuracy gap with general-purpose coding agents at lower per-page cost, and the three leading systems collectively succeed on \topThreeUnionPerfect{}\%, suggesting room for improvement by combining their strengths. Our trace analysis shows that coding agents can perform sophisticated visual measurements and computations, yet still fail through incorrect interpretation of document-specific rules. These results motivate extraction systems that can reliably identify and carry those rules through every step of a task, alongside benchmarks that measure complete document extraction rather than average value accuracy alone.

\paragraph{Limitations.}
\label{sec:limitations}
Pilot performance influenced document selection, and some of those same extractors were then evaluated on the same dataset. Despite being composed of realistic elements, challenge-dense documents do not represent ordinary workloads. Diversity was not measured, nor was human extraction performance studied. We report one result per system and document, so variability was not studied.

%% file: sections/appendix.tex
\appendix

\section{Dataset Details}
\label{app:dataset}
\input{sections/appendix/dataset}

\clearpage
\section{Generator Details}
\label{app:generator}
\input{sections/appendix/generator}

\section{Evaluation Protocol}
\label{app:protocol}
\input{sections/appendix/protocol}

\section{Detailed Results}
\label{app:results}
\input{sections/appendix/results}

\clearpage
\section{Failure Cases}
\label{app:failure-cases}
\input{sections/appendix/failures}

\clearpage
\section{Additional Analyses}
\label{app:additional-analysis}
\input{sections/appendix/additional_analysis}

%% file: sections/appendix/dataset.tex
\FloatBarrier
\subsection{Gallery}
\label{app:gallery}

\begin{figure}[H]
\centering
\setlength{\tabcolsep}{2pt}
\renewcommand{\arraystretch}{0.9}
\begin{minipage}[t]{0.19\textwidth}\centering\includegraphics[width=\linewidth]{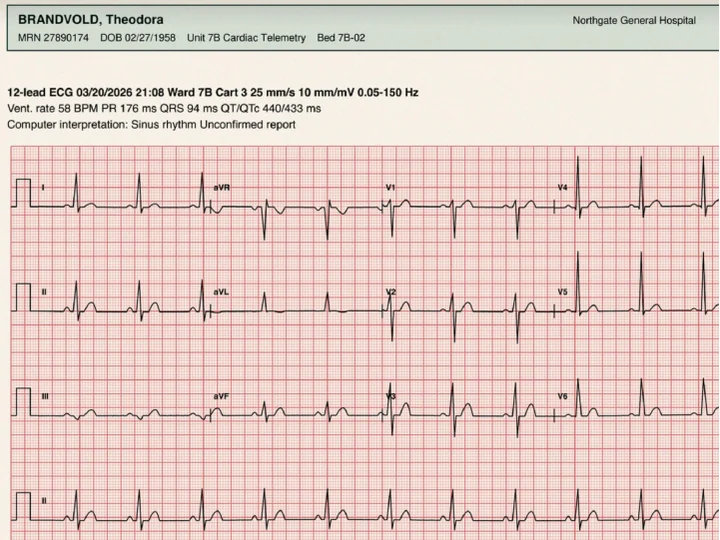}\\[-1pt]{\scriptsize ECG rhythm strips}\end{minipage}\hfill
\begin{minipage}[t]{0.19\textwidth}\centering\includegraphics[width=\linewidth]{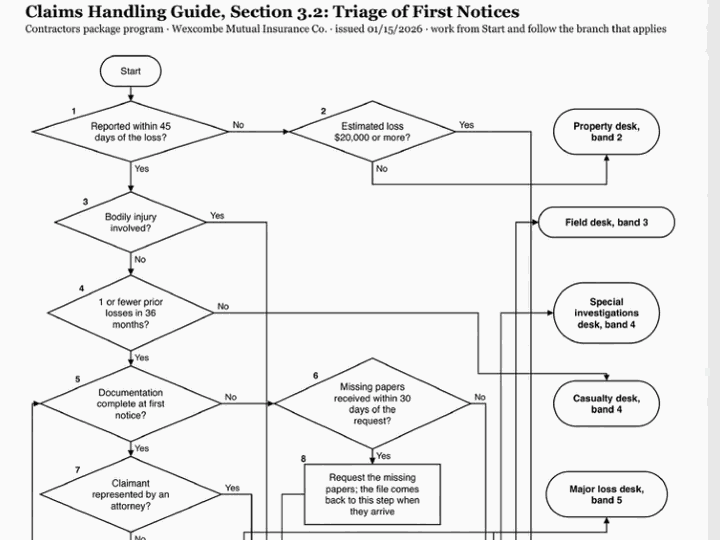}\\[-1pt]{\scriptsize Decision flowchart}\end{minipage}\hfill
\begin{minipage}[t]{0.19\textwidth}\centering\includegraphics[width=\linewidth]{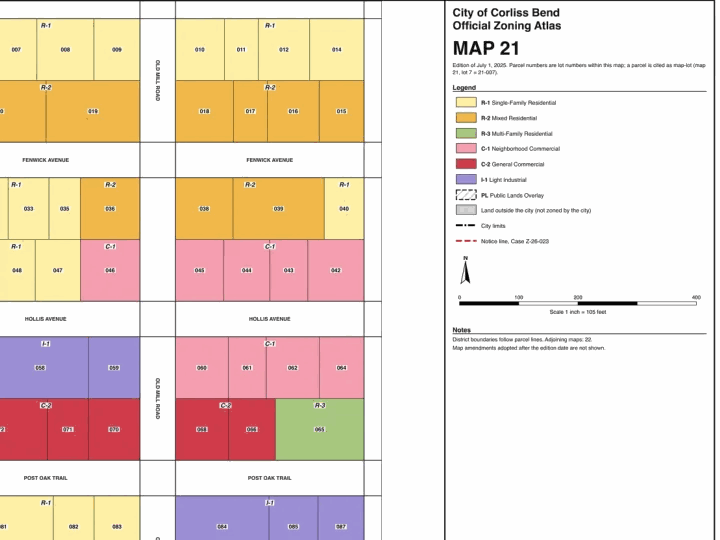}\\[-1pt]{\scriptsize Zoning atlas map}\end{minipage}\hfill
\begin{minipage}[t]{0.19\textwidth}\centering\includegraphics[width=\linewidth]{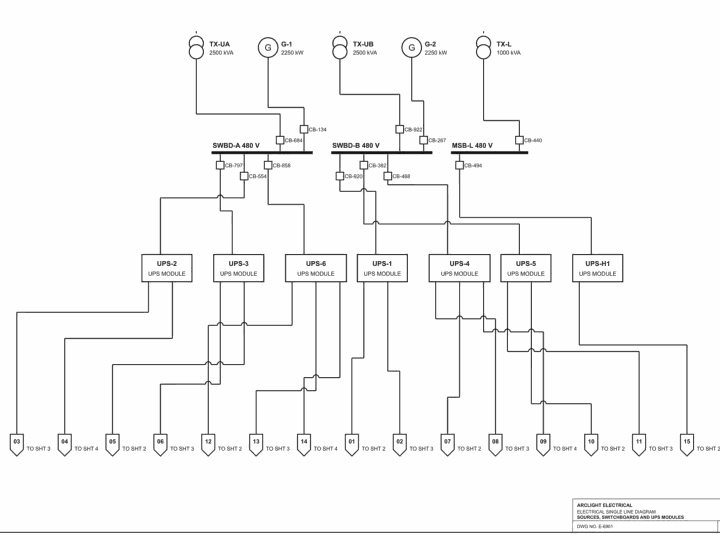}\\[-1pt]{\scriptsize Single-line power diagram}\end{minipage}\hfill
\begin{minipage}[t]{0.19\textwidth}\centering\includegraphics[width=\linewidth]{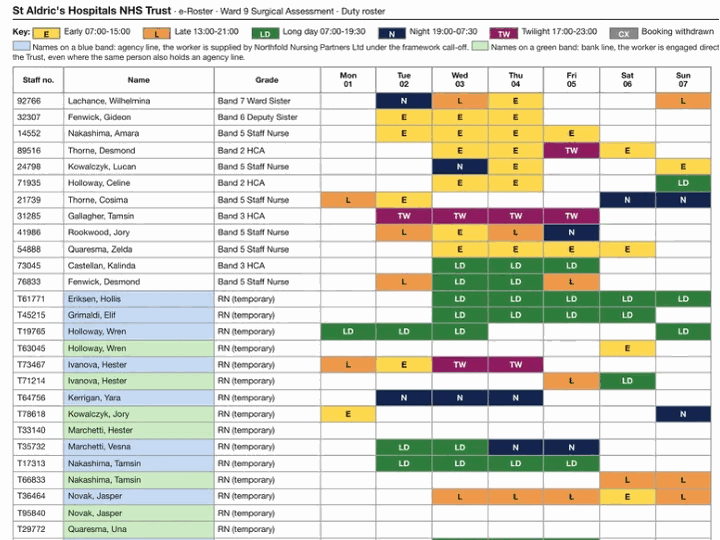}\\[-1pt]{\scriptsize Colour-coded shift roster}\end{minipage}\\[4pt]
\begin{minipage}[t]{0.19\textwidth}\centering\includegraphics[width=\linewidth]{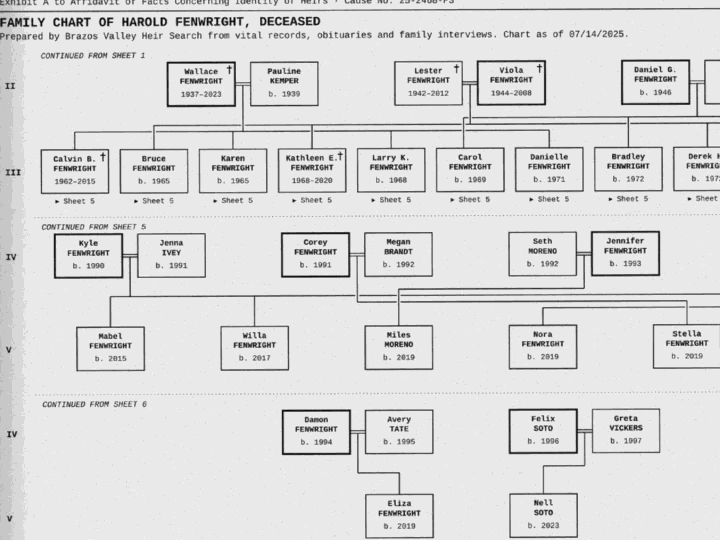}\\[-1pt]{\scriptsize Probate family tree}\end{minipage}\hfill
\begin{minipage}[t]{0.19\textwidth}\centering\includegraphics[width=\linewidth]{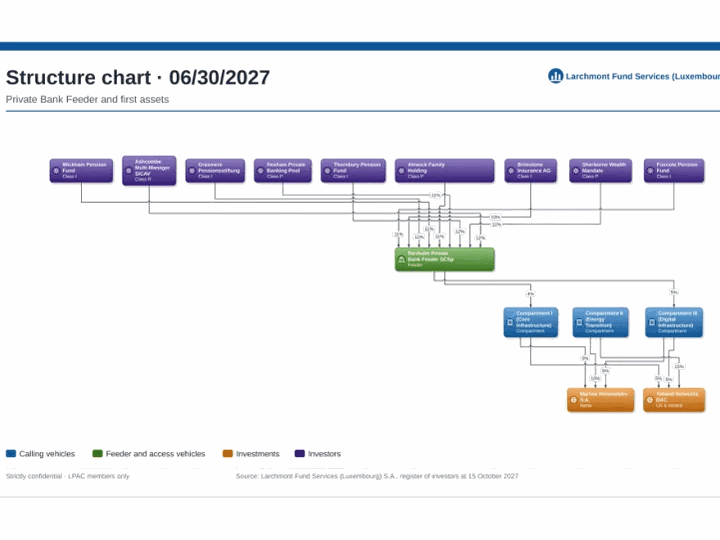}\\[-1pt]{\scriptsize Fund structure chart}\end{minipage}\hfill
\begin{minipage}[t]{0.19\textwidth}\centering\includegraphics[width=\linewidth]{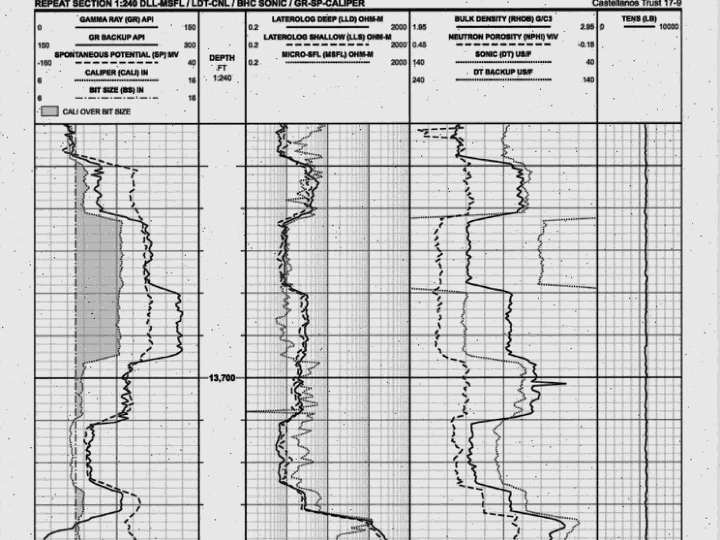}\\[-1pt]{\scriptsize Wireline well-log tracks}\end{minipage}\hfill
\begin{minipage}[t]{0.19\textwidth}\centering\includegraphics[width=\linewidth]{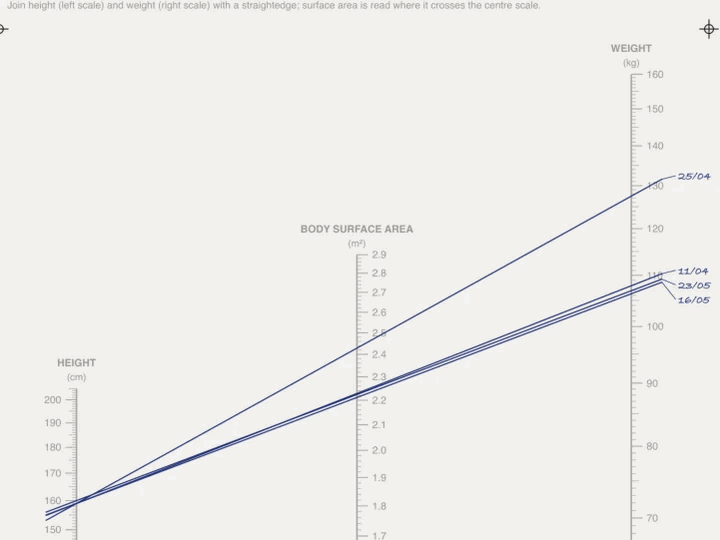}\\[-1pt]{\scriptsize Nomogram with straightedge}\end{minipage}\hfill
\begin{minipage}[t]{0.19\textwidth}\centering\includegraphics[width=\linewidth]{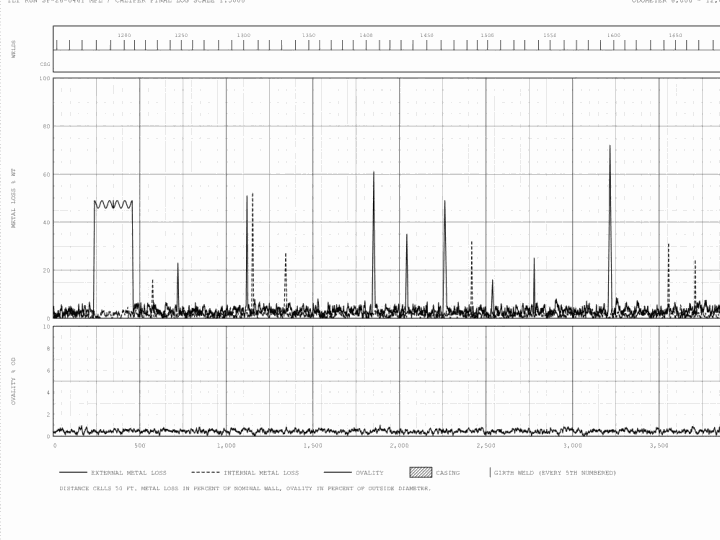}\\[-1pt]{\scriptsize Pipeline inspection strip log}\end{minipage}\\[4pt]
\begin{minipage}[t]{0.19\textwidth}\centering\includegraphics[width=\linewidth]{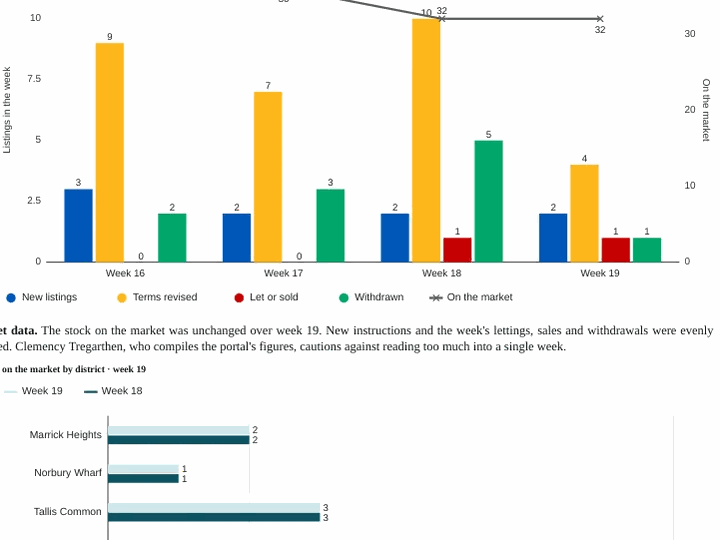}\\[-1pt]{\scriptsize Bars-and-line combo chart}\end{minipage}\hfill
\begin{minipage}[t]{0.19\textwidth}\centering\includegraphics[width=\linewidth]{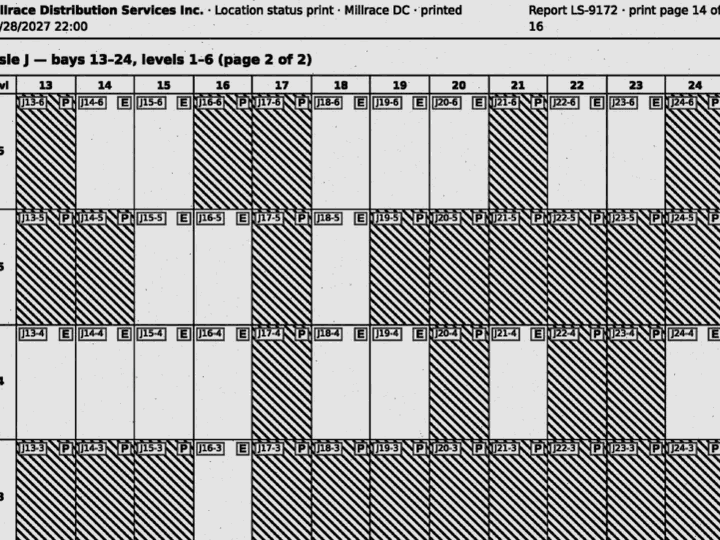}\\[-1pt]{\scriptsize Warehouse slot hatch map}\end{minipage}\hfill
\begin{minipage}[t]{0.19\textwidth}\centering\includegraphics[width=\linewidth]{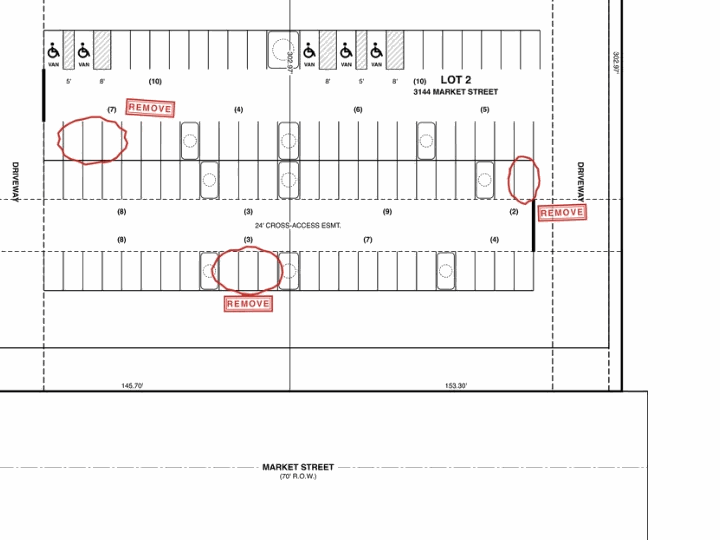}\\[-1pt]{\scriptsize Site plan with review marks}\end{minipage}\hfill
\begin{minipage}[t]{0.19\textwidth}\centering\includegraphics[width=\linewidth]{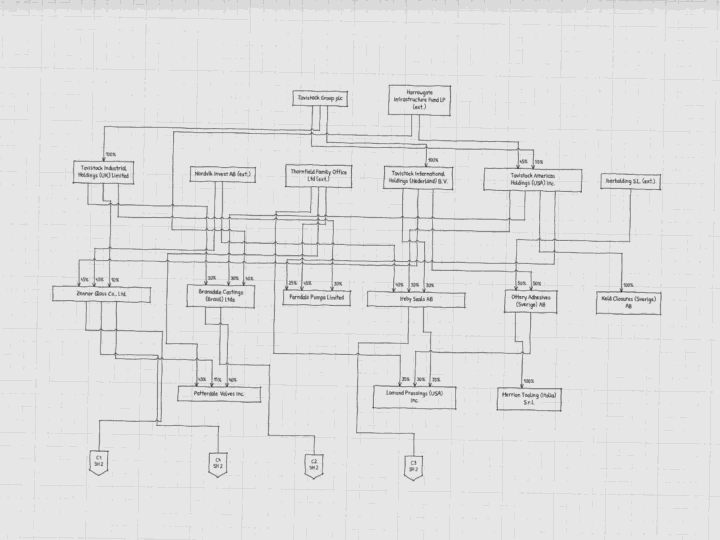}\\[-1pt]{\scriptsize Hand-drawn ownership network}\end{minipage}\hfill
\begin{minipage}[t]{0.19\textwidth}\centering\includegraphics[width=\linewidth]{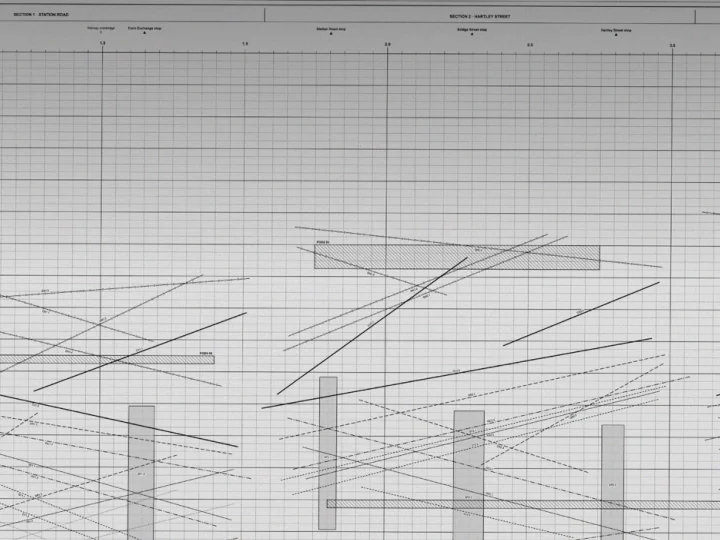}\\[-1pt]{\scriptsize Time-location chart}\end{minipage}\\[4pt]
\begin{minipage}[t]{0.19\textwidth}\centering\includegraphics[width=\linewidth]{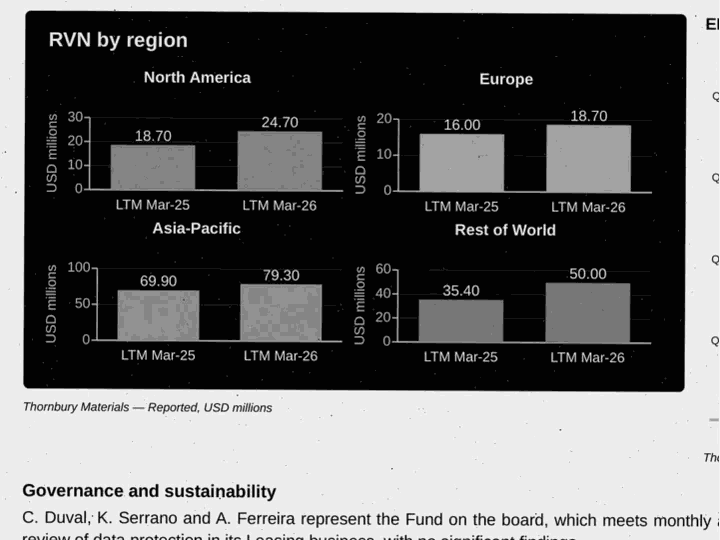}\\[-1pt]{\scriptsize Dark-theme bar panel}\end{minipage}\hfill
\begin{minipage}[t]{0.19\textwidth}\centering\includegraphics[width=\linewidth]{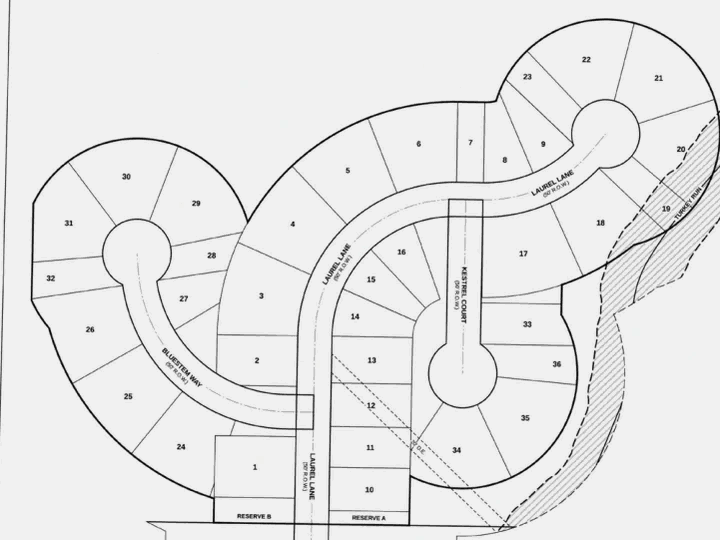}\\[-1pt]{\scriptsize Subdivision plat}\end{minipage}\hfill
\begin{minipage}[t]{0.19\textwidth}\centering\includegraphics[width=\linewidth]{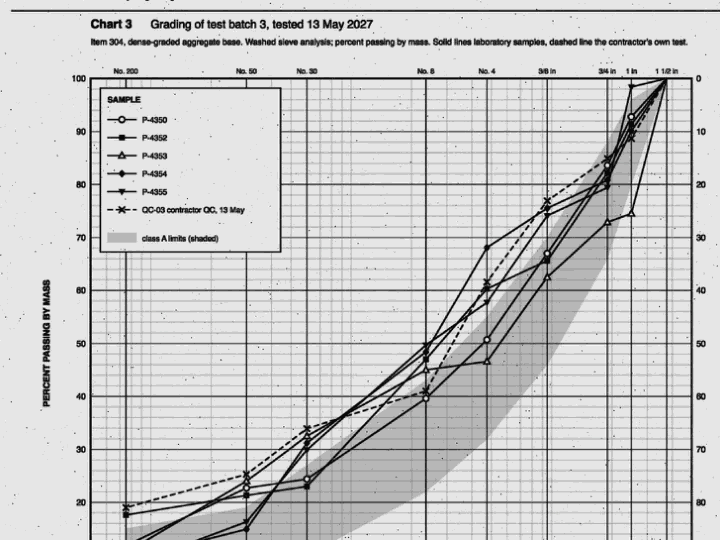}\\[-1pt]{\scriptsize Sieve grading curves}\end{minipage}\hfill
\begin{minipage}[t]{0.19\textwidth}\centering\includegraphics[width=\linewidth]{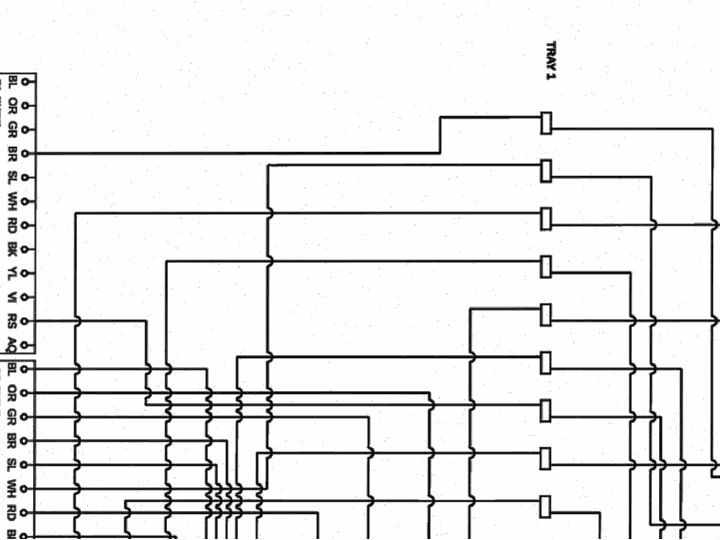}\\[-1pt]{\scriptsize Fibre splice diagram}\end{minipage}\hfill
\begin{minipage}[t]{0.19\textwidth}\centering\includegraphics[width=\linewidth]{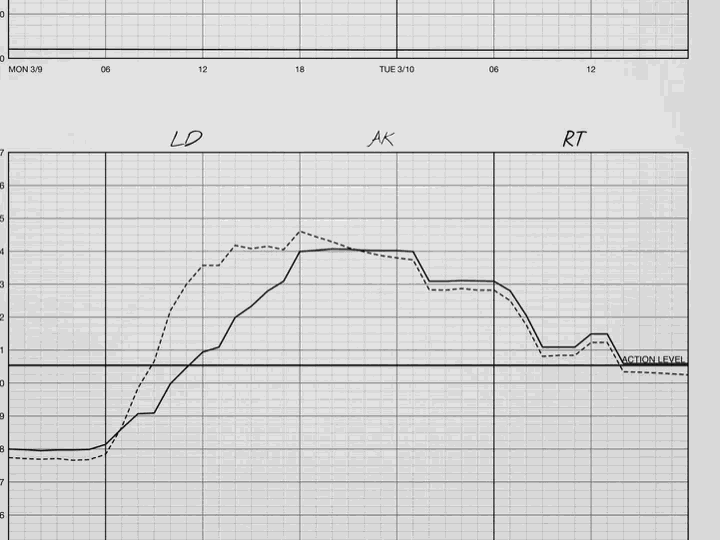}\\[-1pt]{\scriptsize Reservoir hydrograph strips}\end{minipage}
\caption{Pages from twenty documents of ExtractBench Pro, chosen to show the drawn and charted formats.}
\label{fig:gallery}
\end{figure}

The gallery illustrates the range of drawing and chart formats in the generated documents (\Cref{fig:gallery}).

\FloatBarrier
\subsection{Examples}
\label{app:cards}

\newpage
\subsubsection{Interpretation rules}

\begin{figure}[H]
\centering
\includegraphics{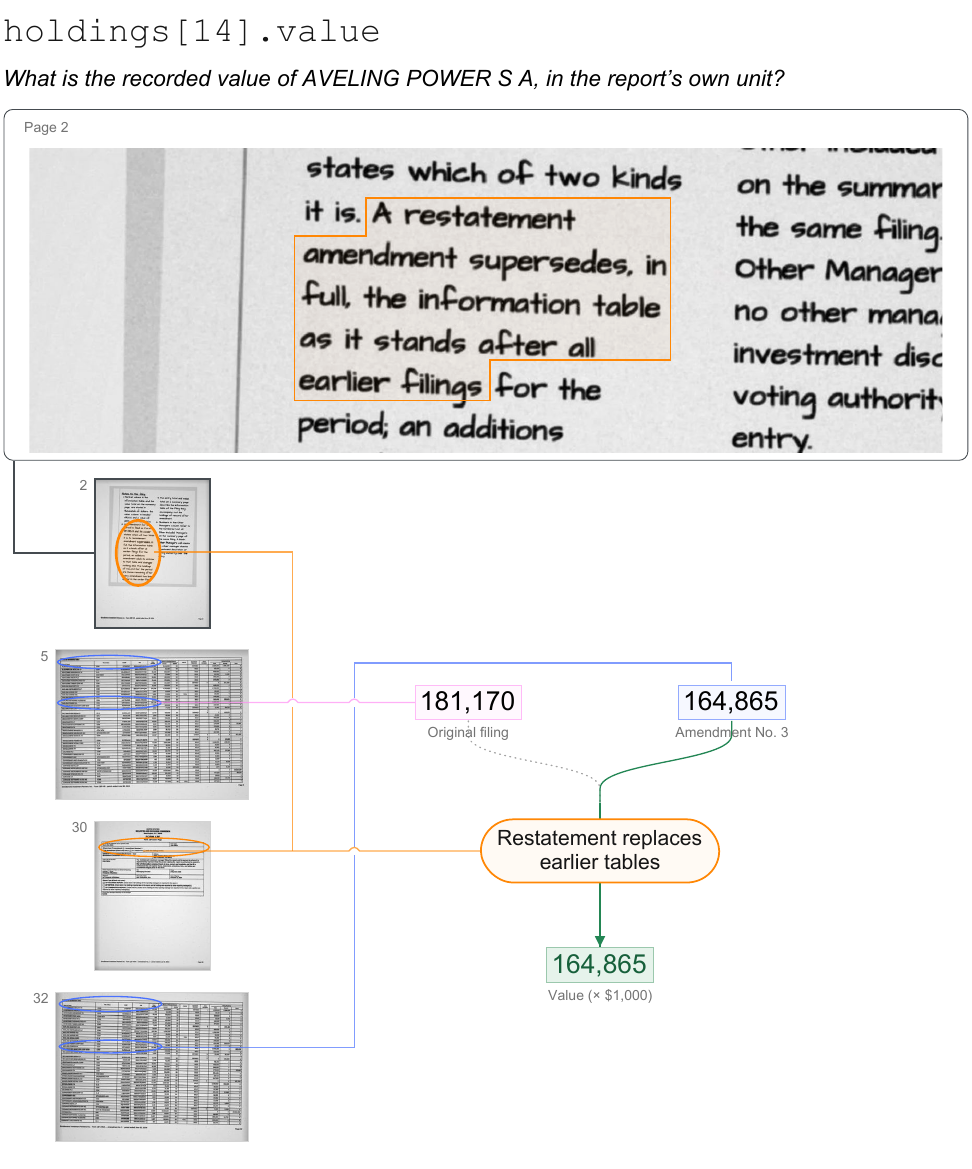}
\caption{An example reasoning flow for extracting a value from a document in the Interpretation rules challenge group.}
\label{fig:card-conventions}
\end{figure}

\begin{samepage}
The 13F generator's data model contains an original holdings report and subsequent filings, each with its own holdings table and amendment type. The generator renders these filings and computes the effective holdings by applying the amendments in order. An extractor must locate AVELING POWER S A, identified by CUSIP WPUI56AD1, which appears in four tables: the original holdings table on page 5, and three amendment tables on pages 14, 23, and 32. To extract the value, the extractor must read Note 2 on page 2, which says that an amendment of type restatement replaces the earlier tables in full. The cover of amendment No. 3 on page 30 has the restatement box checked, so its corresponding table on page 32 takes precedence. Therefore, the extractor should read the value 164,865 from that table.

If the last restatement is ignored and the original table is used, the highlighted value becomes 181,170. The same missed instruction can retain many other superseded holdings and their associated fields from the page 5 table, so this mistake can result in a large number of incorrect extractions: 308 of 3,986 fields, or a penalty of about 7.7 percentage points of lost recall.
\end{samepage}

\FloatBarrier
\newpage
\subsubsection{Custom terminology}

\begin{figure}[H]
\centering
\includegraphics{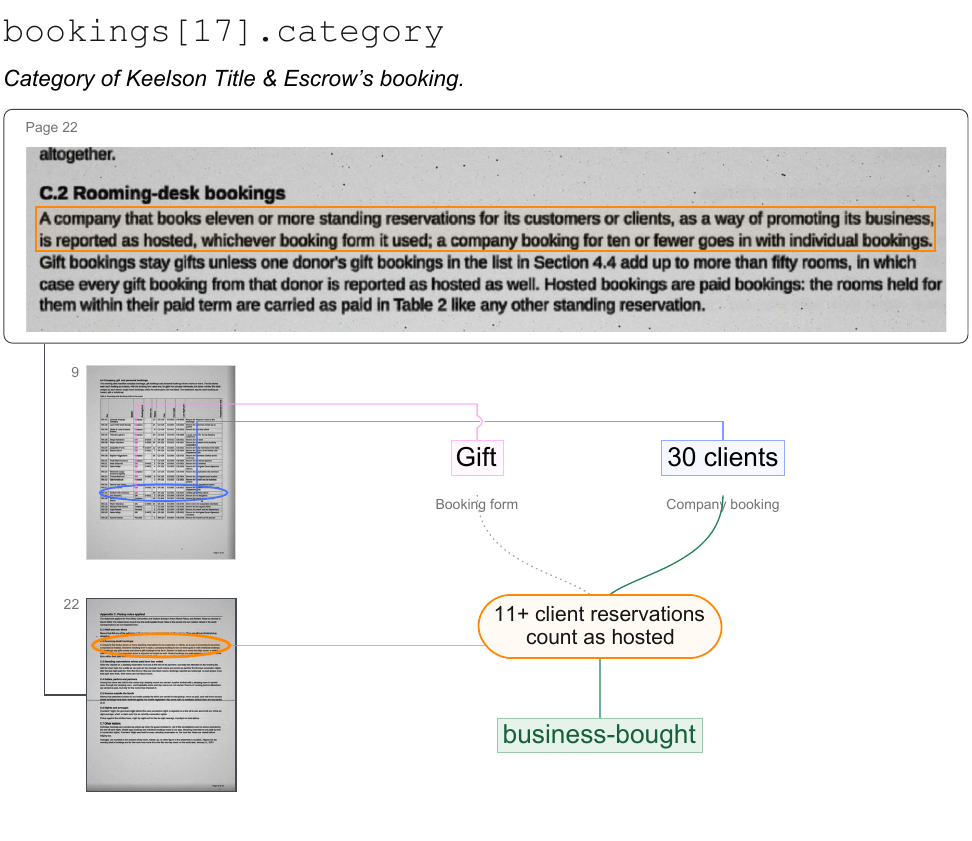}
\caption{An example reasoning flow for extracting a value from a document in the Custom terminology challenge group.}
\label{fig:card-vocabulary}
\end{figure}

\begin{samepage}
The hotel statement's data model records each booking's form, company status, and number of rooms. In this statement, ``hosted'' has a special meaning supplied by rule C.2 on page 22 (a company booking for 11 or more clients is considered ``hosted'' no matter what form was used) and by the schema (a ``hosted'' booking is ``business-bought''). Therefore, when the extractor processes booking RB-118 on page 9, where Keelson Title \& Escrow has used a Gift form for 30 client reservations, rule C.2 applies and causes those bookings to be reclassified as business-bought.

If Rule C.2 is ignored and the booking form is used to determine the category, RB-118 is incorrectly classified under the ``gift'' category. The same missed rule can lead to errors on 27 of the 168 expected fields in the document, resulting in a recall loss of about 16.1 percentage points.
\end{samepage}

\FloatBarrier
\newpage
\subsubsection{Conditional aggregation}

\begin{figure}[H]
\centering
\includegraphics{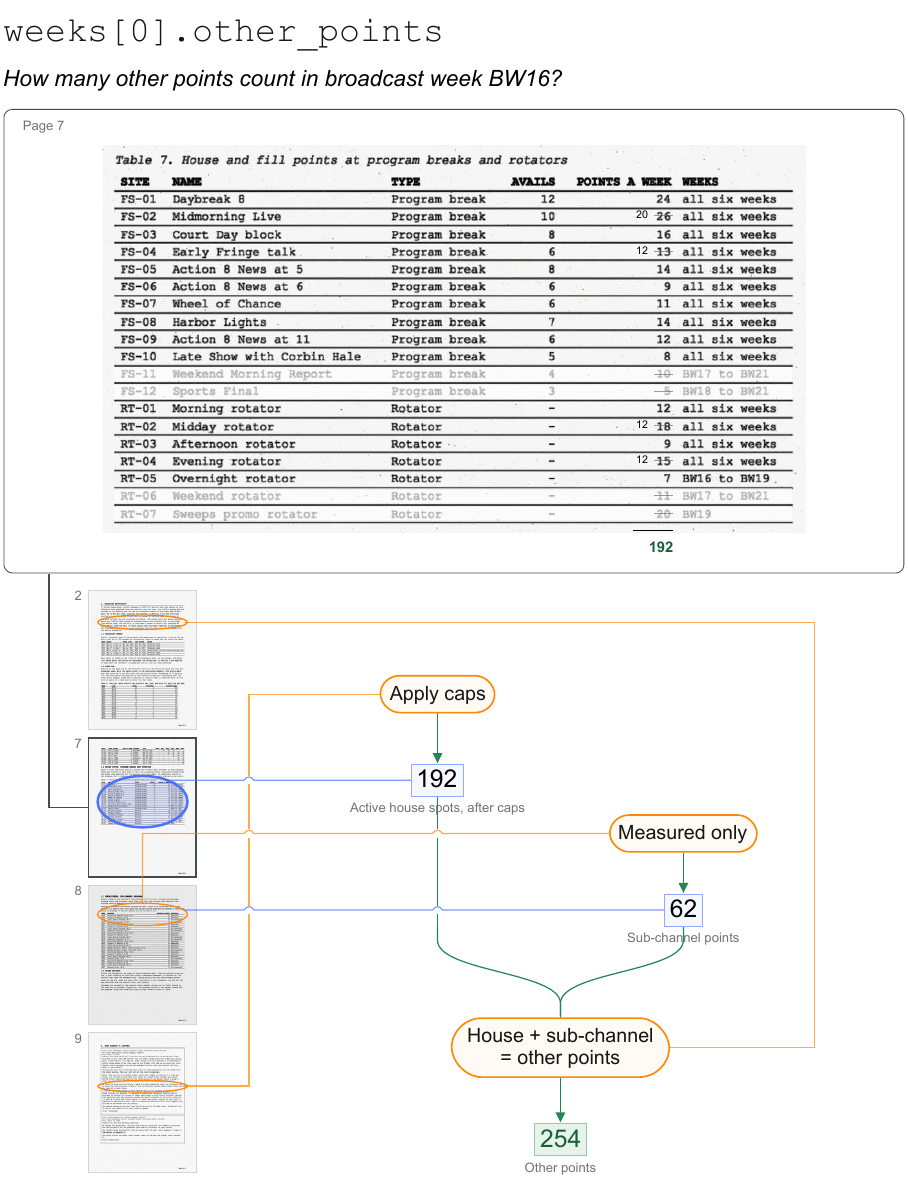}
\caption{An example reasoning flow for extracting a value from a document in the Conditional aggregation challenge group.}
\label{fig:card-counting}
\end{figure}

\begin{samepage}
The broadcast report's data model records weekly audience-rating points, active weeks, commercial avails (advertising slots), and whether sub-channel programs are measured. The generator renders these records in separate tables and computes the expected totals by applying the eligibility rules and caps. To extract ``other points'' for BW16, an extractor must read the definition on page 2, which states that the other points include both eligible house spots and measured sub-channel points. First, for house spots, the house-spot table on page 7 lists the raw points and active weeks for each program. However, before aggregating them, the extractor must select the rows active in BW16 and apply the caps on page 9. The caps allow two points per commercial avail and twelve points per rotator. For example, FS-02 has 26 raw points and ten avails, so it contributes only 20 points. After applying those caps and filtering to active rows, the extractor obtains a total of 192 house points. Second, the extractor must read the sub-channel table on page 8 and include only measured programs, which contribute 27 and 35 points. The answer is therefore 192 + 27 + 35 = 254 other points.

If the cap is missed for FS-02, its contribution becomes 26 rather than 20 points, making the entire total incorrect. Missing caps for even just that one row would make 21 aggregate fields incorrect out of the document's 35 expected fields, producing a recall loss of 60 percentage points.
\end{samepage}

\FloatBarrier
\newpage
\subsubsection{Relational diagrams}

\begin{figure}[H]
\centering
\includegraphics{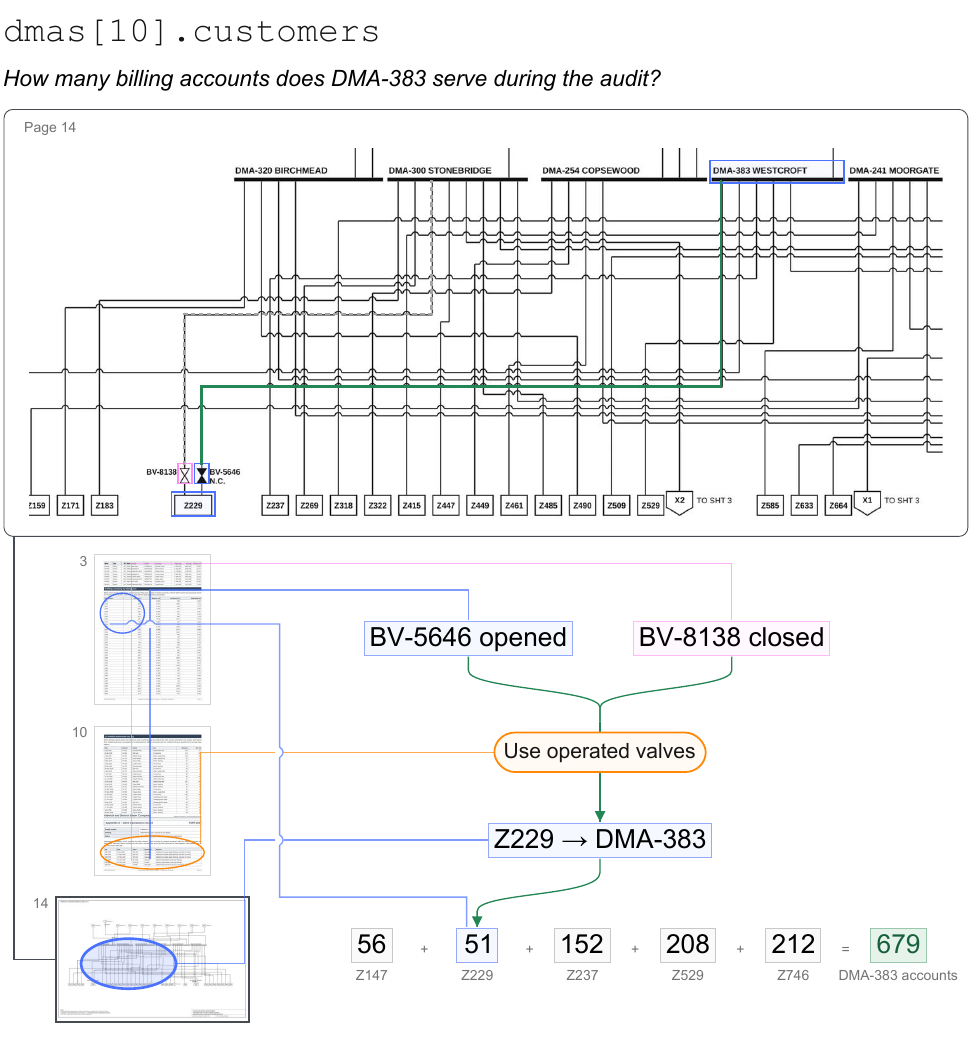}
\caption{An example reasoning flow for extracting a value from a document in the Relational diagrams challenge group.}
\label{fig:card-drawn}
\end{figure}

\begin{samepage}
The water-audit generator's data model contains a pipe network, valve states, billing zones, account counts, and billed volumes. The generator renders the network as a schematic, prints subsequent valve operations separately, and computes the expected aggregates after applying those operations. An extractor must locate zone Z229, a node on the schematic on page 14, and trace its two connecting lines. One passes through valve BV-8138, drawn open, to DMA-300. The other passes through valve BV-5646, drawn normally closed, to DMA-383. The schematic therefore initially places Z229 in DMA-300. However, the operations record on page 10 says that JOB-6685 opened BV-5646 and closed BV-8138 before the audit period. Applying that change moves Z229 into DMA-383. The extractor must then look up Z229 in the billing table on page 3, which gives 51 accounts. Repeating this for 4 other connected zones, we find that they contribute 56, 152, 208, and 212 additional accounts. Adding Z229 to those 4 zones gives the requested count of 679 accounts for DMA-383.

If the valve operations record is ignored, Z229 remains assigned to DMA-300, and DMA-383 is reported as serving 628 rather than 679 accounts. Missing this operation alone makes ten expected fields incorrect across the two districts. Ignoring the entire valve operations record also misses another zone transfer and a change in the pumping stations supplying the network. Together, these mistakes make 21 of the document's 129 expected fields incorrect, producing a recall loss of about 16.3 percentage points.
\end{samepage}

\FloatBarrier
\newpage
\subsubsection{Charts and maps}

\begin{figure}[H]
\centering
\includegraphics{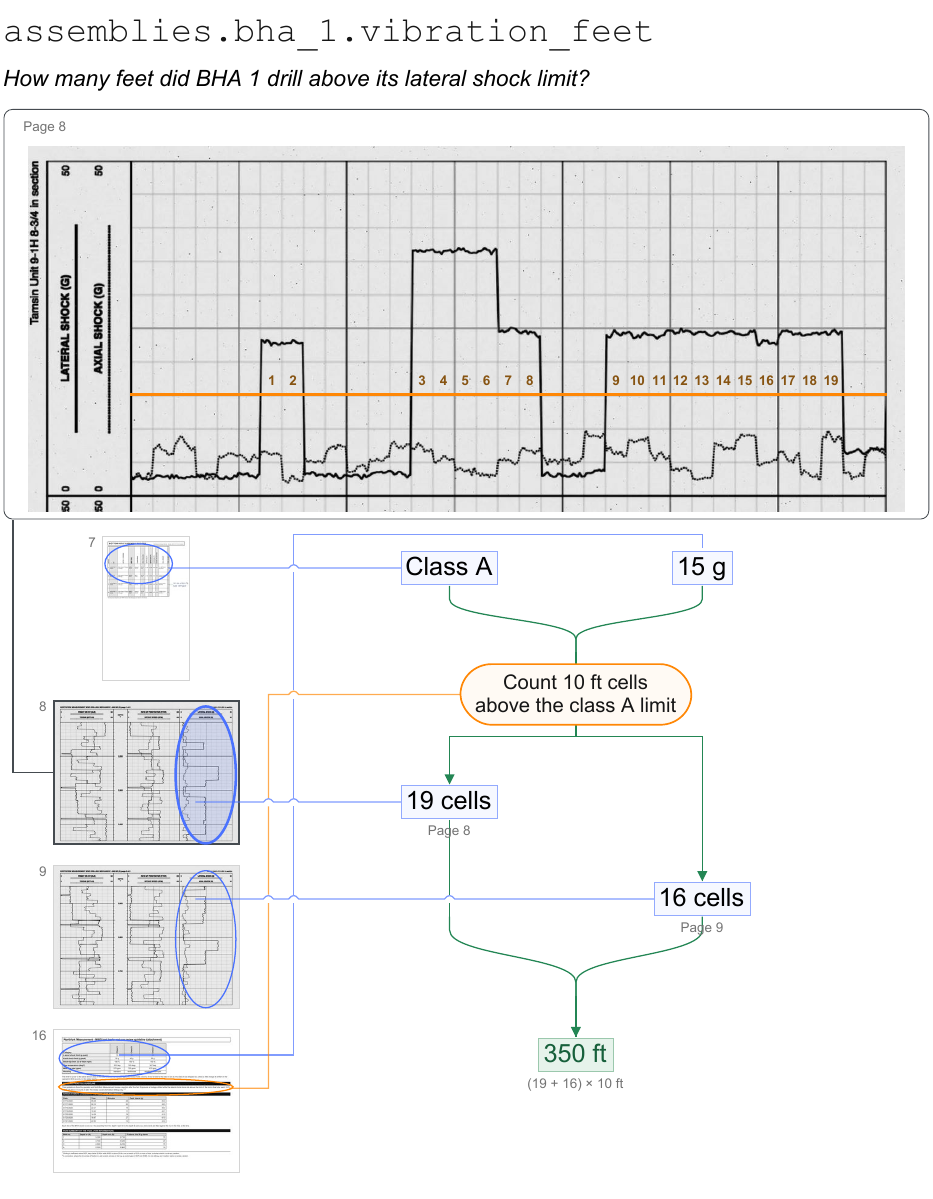}
\caption{An example reasoning flow for extracting a value from a document in the Charts and maps challenge group.}
\label{fig:card-charts}
\end{figure}

\begin{samepage}
The drilling-report generator constructs depth-indexed mechanics curves and tool runs with class-specific operating limits. It renders the curves on mechanics-log sheets and prints the tool classes and limits elsewhere in the report. To extract this value, an extractor must first visit page 7; a table lookup on that page finds that for BHA 1, the Class A limit is 15 g. On the two illustrated sheets (pages 8 and 9), there are 19 and 16 cells above 15 g, respectively. At 10 feet per cell, those 35 cells give an answer of 350 feet.

The report has a generic 30 g alarm; using that limit instead of the BHA 1 Class A limit results in a count of only 7 cells (4 and 3 cells on pages 8 and 9, respectively), giving 70 feet and undercounting the expected answer by 280 feet.
\end{samepage}

\FloatBarrier
\newpage
\subsubsection{Long single pages}

\begin{figure}[H]
\centering
\includegraphics{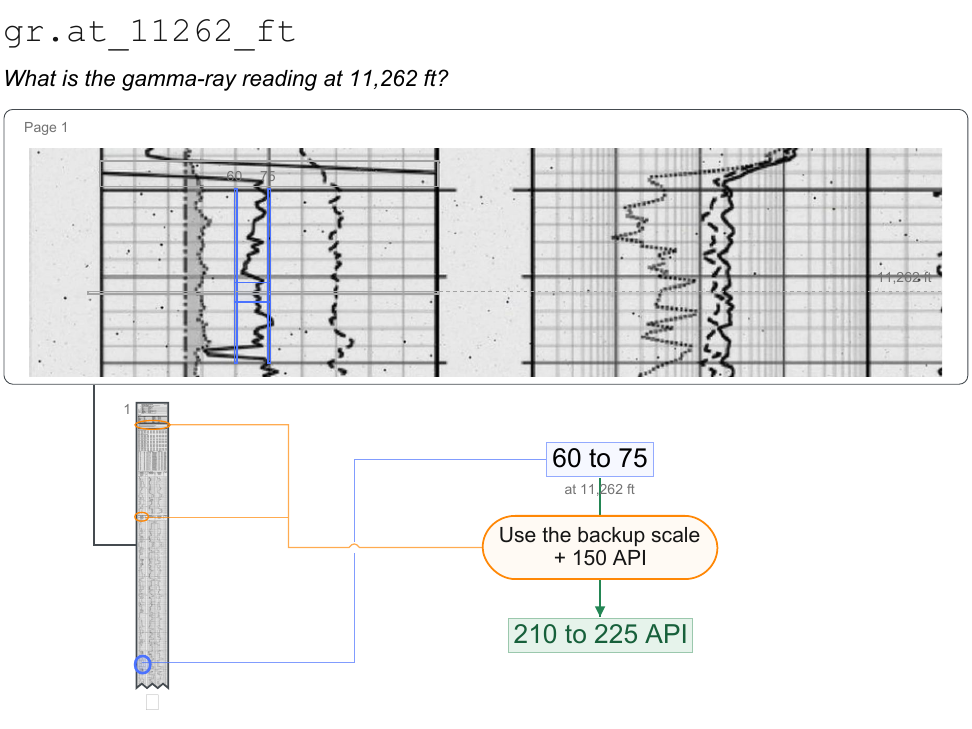}
\caption{An example reasoning flow for extracting a value from a document in the Long single pages challenge group.}
\label{fig:card-long}
\end{figure}

\begin{samepage}
The wireline-log generator's data model contains various curves as a function of depth, and requested depth positions to read values from the curves. As explained at the top of the document, the gamma-ray curve is rendered on a grid with primary (0--150 API) and backup (150--300 API) scales. For example, when the gamma-ray reading goes from 151 to 149, it wraps past the left edge of the grid (exiting the backup scale) and reappears on the right (entering the primary scale). To extract the gamma-ray reading at 11,262 feet, an extractor must first determine which scale applies at that depth. Above 11,262 feet, the trace leaves the right edge of the track and returns from the left, meaning that the 150--300 API backup scale applies. Only then can the position of the trace at 11,262 feet on the depth axis be interpreted. It lies between 60 and 75 API on the primary scale, corresponding to the 210--225 API range on the backup scale.

If the backup-scale instruction is ignored and the printed primary scale is used, the answer becomes 60 to 75 API. Ignoring backup scales in general on this document can produce incorrect intervals and aggregate values for 70 of the 500 requested fields, corresponding to a 14-percentage-point recall loss.
\end{samepage}

%% file: sections/appendix/generator.tex
\subsection{Sources of the Challenges}
\label{app:challenge-sources}

With the goal of identifying challenging document elements and patterns, we asked coding agents to inspect real documents, published definitions, and extraction schemas. The agents recorded where the information needed for an answer appeared, how records were linked across pages, and which rules affected their interpretation. After running a set of pilot extractors on a selection of those documents and schemas, we identified incorrect or missing fields and traced each error back to the relevant pages and schema. This let us describe candidate reasoning paths a reader might have to follow, such as finding an amendment before deciding which version of a table to use. Moreover, we could predict likely errors an extractor might make if it fails to follow the entire reasoning path. For example, when a value is amended, returning the pre-amendment value would suggest that the extractor missed or misapplied the amendment.

Targeting those candidate reasoning paths, we turned those observations into candidate extraction tasks, using coding agents to produce document generators following the same method described in \Cref{sec:agentic-implementation}. Pilot extractors extracted each of the candidate documents, and the coding agent's goal was to make the pilot extractor fail. After enough attempts, if a likely failure could not be induced in enough extractors, then the candidate reasoning path was rejected as not challenging enough.

At the same time, towards visual fidelity, agents gathered examples of layouts, tables, and styles of the reference documents, so we could reproduce similar diversity later in our synthetic documents. We ran pilot extractors, including coding agents and specialized extraction APIs, on the generated PDFs and their schemas. We compared their outputs with the expected answers and inspected errors on fields that depended on the element. Where possible, we computed the answer under a simpler reading and checked whether it explained the observed errors. We also checked for ambiguous instructions, wrong annotations, and rendering defects. The pilot results helped us decide which designs to keep and which to revise and test again.

\subsection{Challenging Elements}
\label{sec:challenges}

The result of the investigation in Appendix~\ref{app:challenge-sources} identified the following collection of elements and features which were most successful at inducing extraction difficulty. The elements and transformations below can be combined in a single document and reused across challenge groups.

Some elements create difficulty by inducing multi-hop reasoning:

\begin{itemize}[leftmargin=1.2em,itemsep=2pt]
\item \textbf{Rules.} A rule tells the reader how to interpret context elsewhere in the document. In our reference documents, they are found in the form of legends, units, exclusions, and amendments, and so on. Difficulty can be increased by placing them in faraway sections during document assembly.
\item \textbf{Jargon.} Document-specific definitions can override familiar meanings of terms, or clarify otherwise vague words. For example, in one grain trade rule, ``prompt'' shipment means within ten calendar days. Useful terminology is found by searching domain-specific terms, or common patterns of internal organizational jargon. To verify the difficulty, we wrote short questions about candidate terms and asked models to answer them without the governing definitions. Then, to exploit that difficulty, we created rules that would force a model to use an interpretation it is not used to.
\item \textbf{Aggregation.} Rather than copying the value from a page, aggregation combines multiple values into a single derived value such as by counting or summing them. This is frequently found in real requests encountered by extraction services. It can be made more difficult by setting conditions on which values contribute to counts and sums (e.g., which ballots count as valid). This can be done through rules supplied by the schema or the document.
\end{itemize}

Other elements require careful visual precision to interpret:

\begin{itemize}[leftmargin=1.2em,itemsep=2pt]
\item \textbf{Handwriting.} We use eight open-license handwriting-like fonts to fill in handwritten fields in synthetic table and note templates.
\item \textbf{Charts and maps.} Extracting values from a chart or map requires precisely identifying the position of a mark relative to axis markings. We found many examples of such charts and maps in the reference documents, and developed component implementations to recreate each of them. To stay compatible with the ExtractBench task format, which has a finite set of accepted values per field, we discretize these extractions into a choice between predetermined intervals aligned with axis ticks.
\item \textbf{Relational diagrams.} Many diagrams in real documents are relational diagrams, which use lines (edges) to show how entities (nodes) are connected. Examples of these include family trees, electrical circuits, and water networks. Relationships can be directional, or labelled with various types and quantities. We integrate these into larger multi-hop reasoning chains by introducing inferred relationships, or derived values that are conditional on relationships.
\item \textbf{Extreme aspect ratios.} We take inspiration from document formats designed to be printed as continuous sheets, including wireline logs and railway track charts. The resulting PDFs would be several metres long if printed, so a full-page view can make the contents too small to read, while a crop can lose long-range context like axis scales.
\end{itemize}

Finally, a set of perception-related transformations can be used to create obstacles and distractions that must be handled before the document can be cleanly processed.

\begin{itemize}[leftmargin=1.2em,itemsep=2pt]
\item \textbf{Scan artifacts.} The way that a document is physically materialized and digitized can make it more difficult to visually interpret a page. Faded toner, for example, can make small text hard to distinguish from the background. We recorded skew, fading, and bleed-through while inspecting reference scans. Our capture recipes simulate scanner, photocopy, fax, aging, and phone-photo effects. To ensure that text is still legible, a readability gate checks evidence-text size and contrast and reduces degradation when needed.
\item \textbf{Unreliable scans.} A scanned PDF can contain an invisible OCR layer added by the scanner that differs from the visible page. For example, it might record a zero as the letter O, or vice versa. A system relying on the hidden text can therefore return a wrong value from an otherwise legible page. To simulate unreliable scans, we randomly substitute letters in simulated text layers with commonly mistaken alternatives that have similar glyphs.
\item \textbf{Page arrangement.} In practice, scanned documents are assembled by humans who may introduce duplicated, rotated, misplaced, or unrelated pages by mistake. Sometimes, scanners combine multiple pages together in a single image. A reader that blindly trusts such a PDF to represent a page-by-page structure may produce incorrect extractions. We noted places in reference documents where separate subdocuments were bundled together and where page numbering restarted. Then, we reproduced these features in synthetic documents by sampling random swaps, duplications, insertions, and compositions.
\end{itemize}

\subsection{A Page at Four Stages}

\input{sections/appendix/generation_figure}

\FloatBarrier
\begin{samepage}
\subsection{Validation}
\label{app:checks}

We check that the rendered document supports the expected answers. Text checks compare quoted evidence with its assigned page region. For graphical readings, some checks use specially modified rendering. For example, a second wireline-log render assigns distinct colours to curves so that the requested grid-cell readings can be verified more easily.

When the printed instructions support more than one interpretation of the schema and document, we require the ground truth to accept the answers corresponding to each supported reading. Additional readings were often discovered at the pilot extractor stage of validation. The generator implementations support this by computing answers for all acceptable readings from the same underlying data model.
\end{samepage}

%% file: sections/appendix/generation_figure.tex
\begin{figure}[!htb]
\centering
\begin{subfigure}[t]{0.24\textwidth}\includegraphics[width=\linewidth]{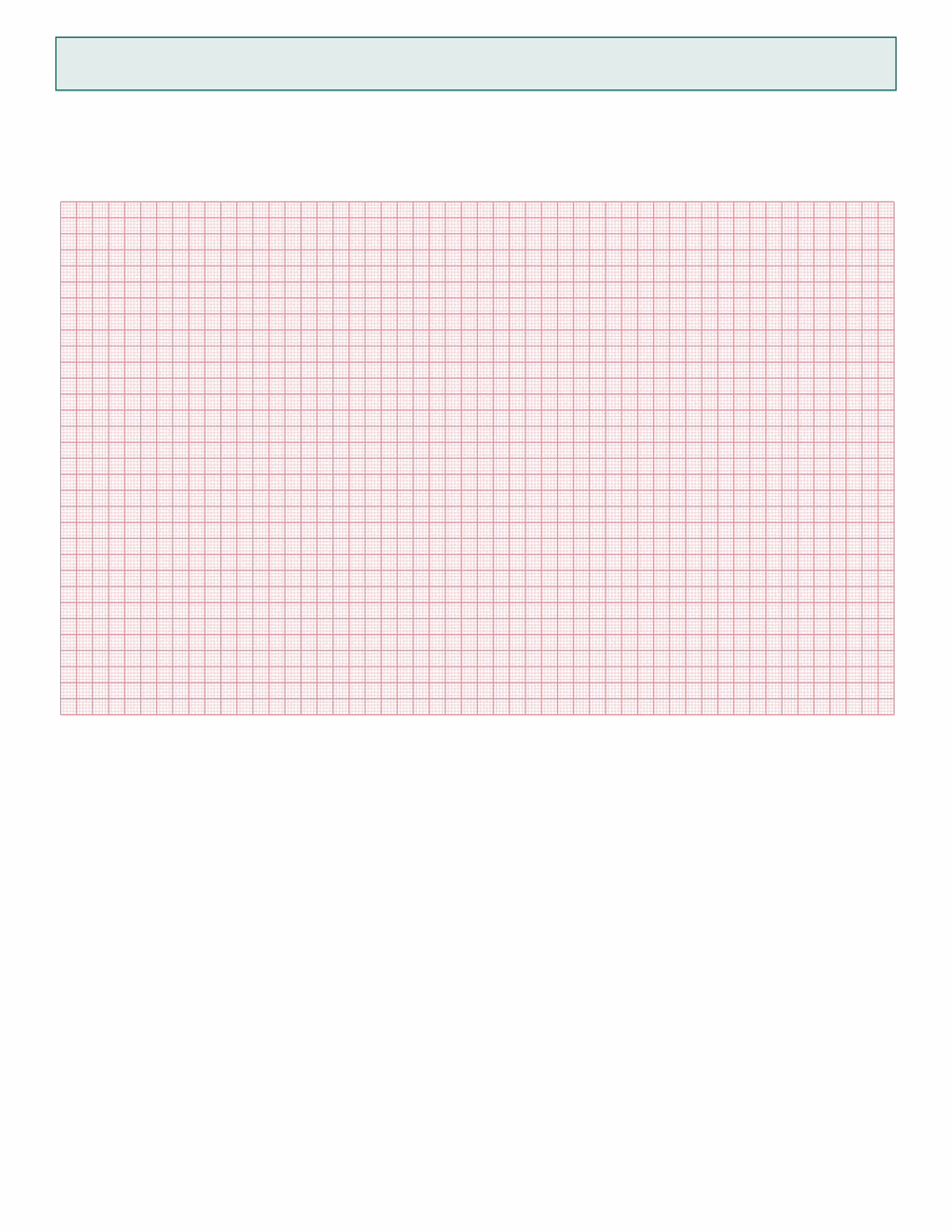}\caption{Page template}\end{subfigure}\hfill
\begin{subfigure}[t]{0.24\textwidth}\includegraphics[width=\linewidth]{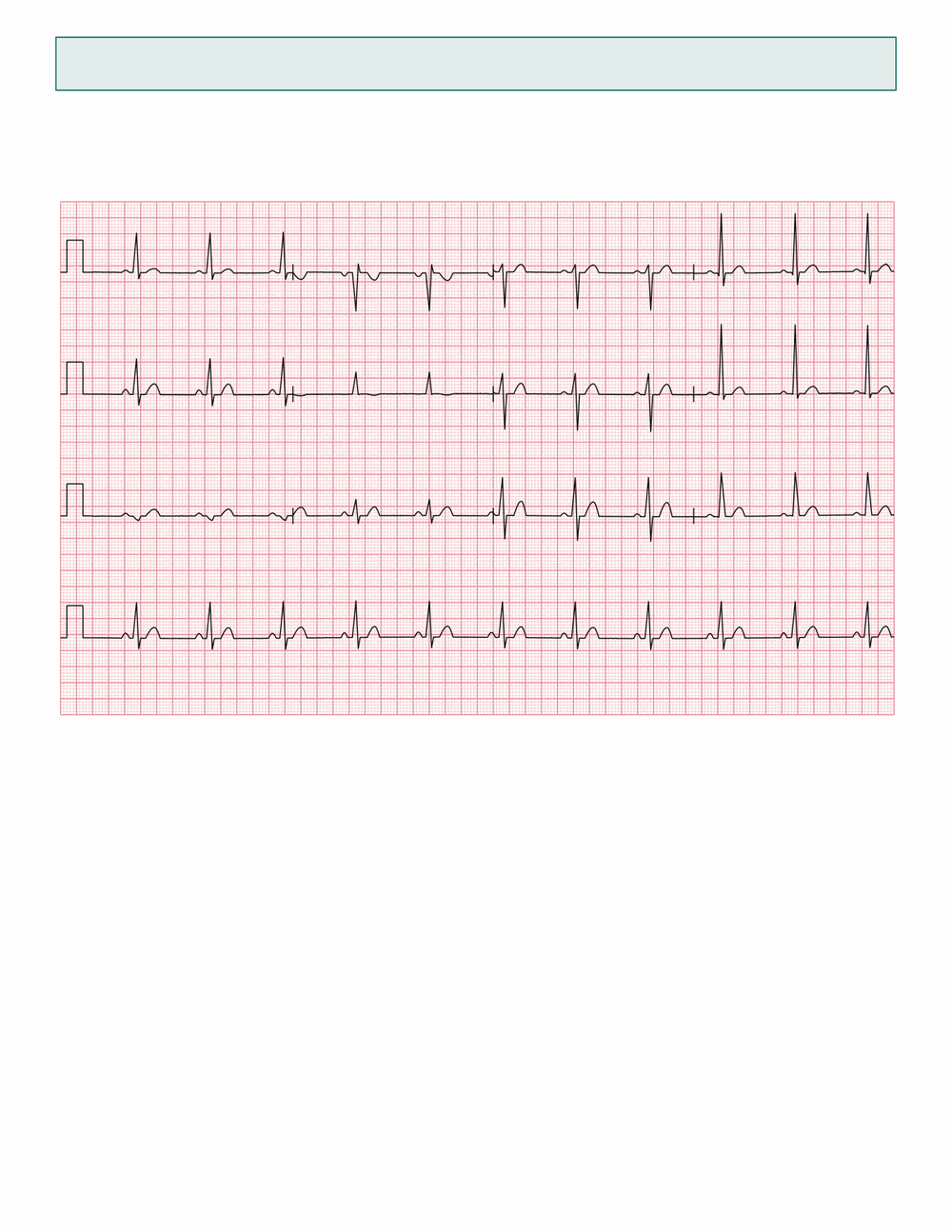}\caption{Drawn components}\end{subfigure}\hfill
\begin{subfigure}[t]{0.24\textwidth}\includegraphics[width=\linewidth]{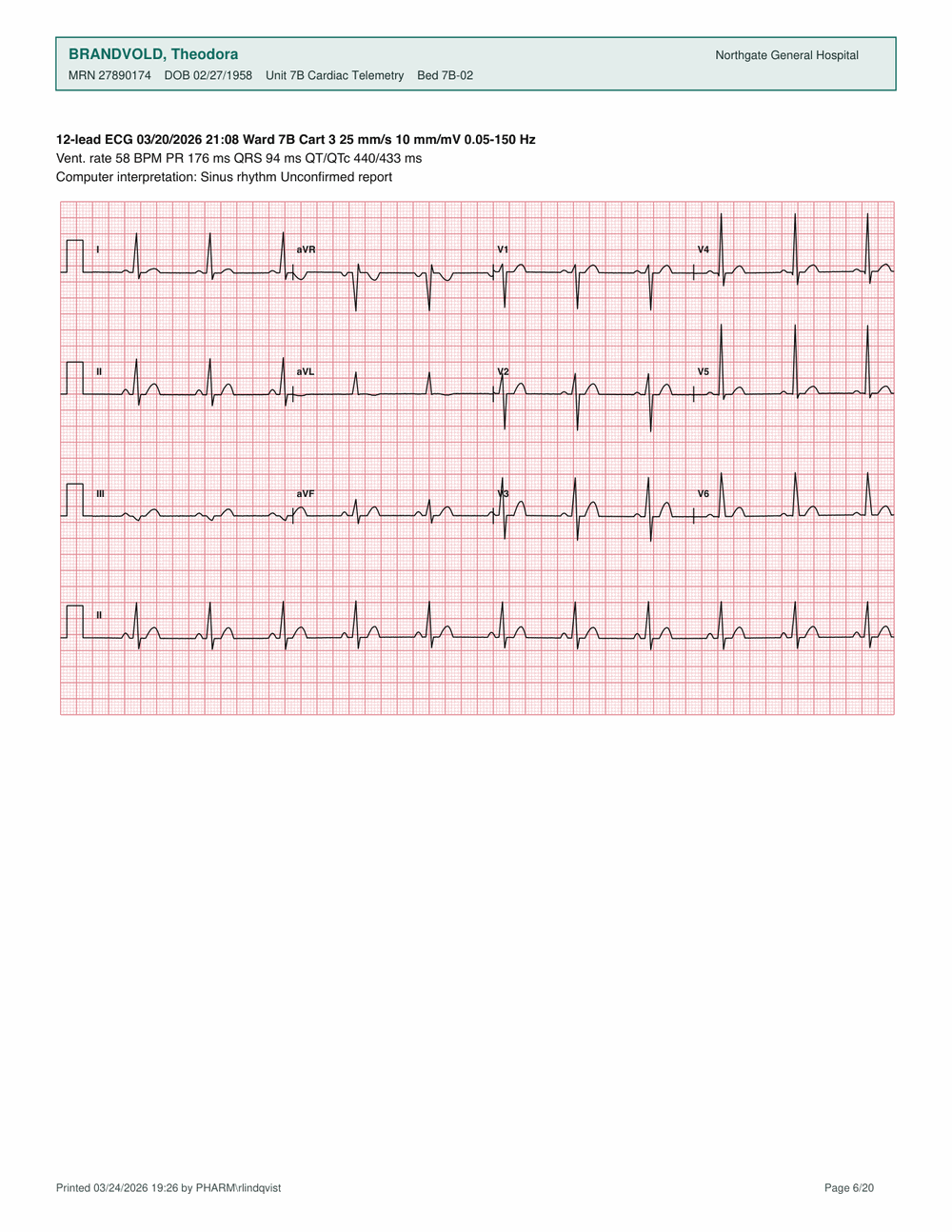}\caption{Clean print}\end{subfigure}\hfill
\begin{subfigure}[t]{0.24\textwidth}\includegraphics[width=\linewidth]{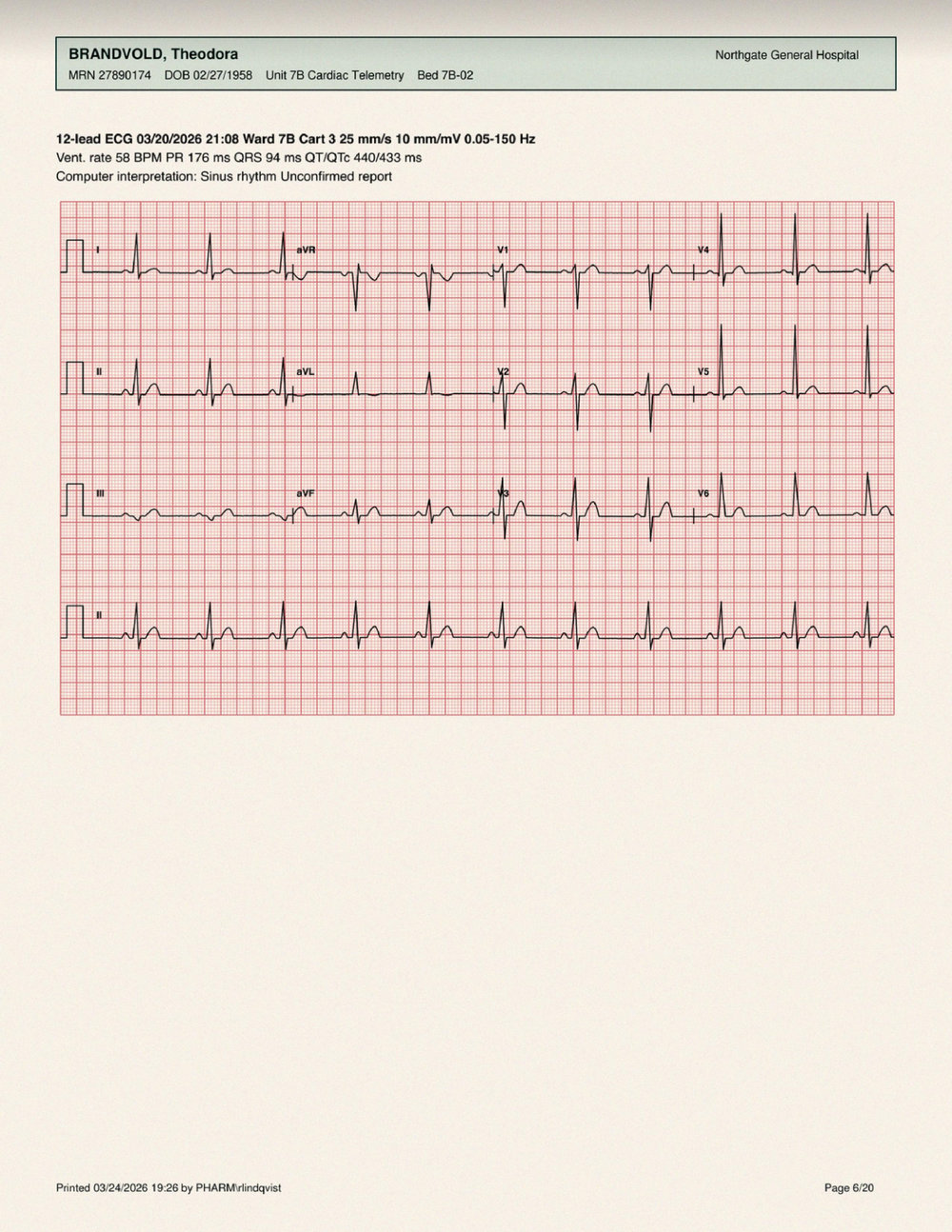}\caption{Shipped page}\end{subfigure}
\caption{Four stages in generating an ECG page for a titration packet. The program lays out the page, draws traces from the data model, adds text, and applies a capture transform. Expected values come from the underlying data. Evidence annotations are added where the task provides a localized source.}
\label{fig:generation}
\end{figure}

%% file: sections/appendix/protocol.tex
\subsection{Systems and Data Versions}

All evaluated systems received the same PDF and schema for each document.
Coding agents ran through vendor command-line tools in an isolated working directory containing those files.
They could use local tools, with network access restricted to the model API, and were prompted to return the page and box supporting each value.
Codex GPT-6 Astra used the highest reasoning setting, GPT-6.1 Sol the lowest, and GPT-6 Luna and GPT-6.1 Terra the medium setting; Muse Code ran Muse Ember 1.0 at the high setting; Gemini CLI ran Argon at medium thinking and Hydrogen at high thinking with vision enabled; Claude Code used medium effort for Opus 5.5, Sonnet 5.5, and Haiku 5.5.
Extraction APIs received the inputs through their public interfaces with evidence enabled where available.
LlamaExtract's Agentic Plus and Agentic tiers, Datalab Accurate (with phrase-level citations), and Reducto Deep use their November 2026 releases; LlamaExtract Agentic Plus Max Effort uses the published November 2026 run; Extend Operator and Pulse Effort use the October 2026 snapshot.

The main results use a consistent three-hour execution limit per document, including retries. The selected runs include the successful repairs of the earlier one-hour Sonnet and Muse timeouts. Saved predictions are evaluated against the same answer key and scorer revision.

Failed and empty runs remain in the accuracy denominators. Completed outputs beyond three hours also score zero. The 30-minute comparison uses the same selected outputs and changes only the scoring cutoff.

\subsection{Cost and Latency}
\label{app:cost}

We use prices per page from the public ExtractBench Pro leaderboard. For Claude Code Haiku and Muse Code Ember, we average the recorded cost per page across documents.

Page count can understate the work required by an oversized chart.
Each long chart sheet counts as one page, so its cost per page equals its cost per sheet.
The most expensive sheet cost Codex GPT-6 Astra \$\axLongMaxCostUsd{}.
Latency is recorded wall-clock time; the reported mean uses runs with a recorded duration, as specified in \Cref{tab:timeouts}.

%% file: sections/appendix/results.tex
\FloatBarrier
\subsection{Grounding by Challenge}

Grounding scores depend on which fields have evidence annotations.
\Cref{tab:grounding} reports these scores with box coverage for each challenge across all \pronsystems{} systems.
Long single pages have complete box coverage but low Grounding F1: the highest coding-agent score is \bestAgentGroundedLongCanon\%, and specialized extraction APIs reach at most \bestApiGroundedLongCanon\%.
For other challenges, the annotated fields may omit computed answers or relationships supported by several regions.

\input{tables/grounding}

\FloatBarrier
\subsection{Latency and Execution Outcomes}
\label{app:time}

\Cref{tab:timeouts} reports runtime and execution outcomes under the three-hour budget.

\input{tables/timeouts}
\input{tables/results_overall_no_limit}

\FloatBarrier
\subsection{Cost by Challenge}

Long single pages account for much higher per-page costs than the other challenges.
On the other five groups, no evaluated configuration exceeds \otherChallengesCostMaxCents{} cents per page.
On long sheets, coding agents spend \agentLongSheetsCostMinCents{} to \longSheetsCostMaxCents{} cents per sheet on average; LlamaExtract Max Effort spends \amCostLong{} cents and Datalab Accurate spends \dlCostLong{} cents.
\Cref{tab:results-overall,tab:cost-by-challenge} give the aggregate and per-challenge costs.

\input{tables/cost_by_challenge}

\FloatBarrier
\subsection{Per-Document Results}
\label{app:hardest}

\begin{figure}[!htb]
\centering
\includegraphics[width=\textwidth]{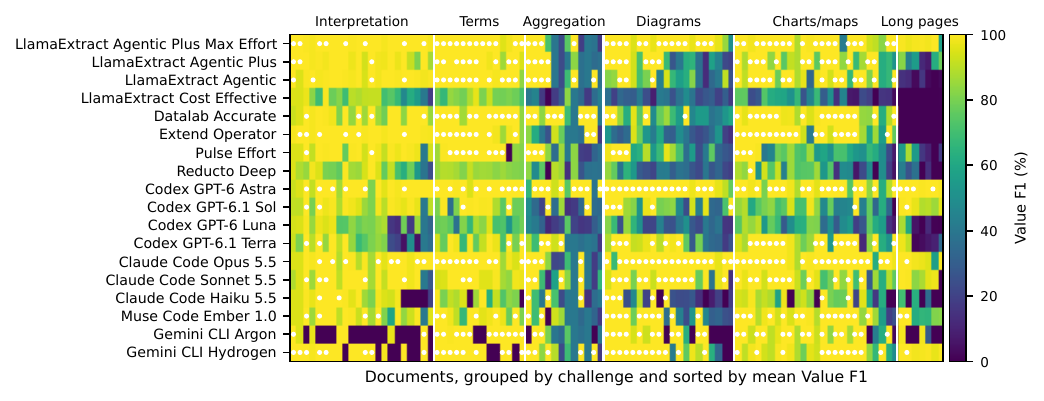}
\caption{Per-document Value F1 under the three-hour budget across the \proPerDocSystems{} systems. Documents are grouped by challenge and ordered by mean score; systems are grouped with extraction APIs first, followed by coding agents. White dots mark documents with Value F1 equal to one.}
\label{fig:score-grid}
\end{figure}

The score matrix shows which documents each system completes or fails.
\Cref{fig:score-grid} groups documents by challenge and marks perfect results with white dots.
Zeros include failed and empty outputs, incorrect extractions, and runs exceeding the three-hour budget.
The best-to-worst gap for each document is computed across the \proPerDocSystems{} systems in the matrix, including those zeros.
A document is counted as unsolved when no system achieves unrounded Value F1 of one.

The hardest documents for the panel are not uniformly hard for every system.
Among the ten documents with the lowest mean Value F1 under the three-hour budget, \bottomTenWithPerfect{} have a perfect result from at least one system, and \bottomTenLongSheets{} are long single pages (\Cref{tab:hardest}).
The lowest best-system score among these ten is \bottomTenMinBest\%, on a long time--location chart.

\input{tables/hardest_documents}

%% file: tables/grounding.tex
\begin{table}[!htb]
\centering
\small
\setlength{\tabcolsep}{3pt}
\renewcommand{\arraystretch}{0.92}
\begin{tabular}{@{}lrrrrrrr@{}}
\toprule
 & \multicolumn{7}{c}{\textbf{Grounding F1}} \\
\cmidrule(l){2-8}
\textbf{System} & All & Interp. & Terms & Aggr. & Diagr. & Charts and maps & Long pages \\
\emph{Boxed values (\%)} & 78 & 77 & 100 & 66 & 20 & 7 & 100 \\
\midrule
\multicolumn{8}{@{}l}{\textit{Coding agents}} \\
\quad Codex GPT-6 Astra & \textbf{72.2} & \textbf{92.2} & \underline{63.5} & \underline{33.3} & \textbf{87.1} & \underline{87.8} & 14.0 \\
\quad Codex GPT-6.1 Sol & 65.0 & 82.0 & 54.2 & 32.2 & \underline{80.3} & 75.7 & 19.2 \\
\quad Codex GPT-6 Luna & 47.6 & 52.7 & 44.1 & 18.2 & 57.0 & 83.0 & 0.0 \\
\quad Codex GPT-6.1 Terra & 60.2 & 63.6 & 53.7 & 29.6 & 79.4 & \textbf{94.1} & 0.9 \\
\quad Claude Code Opus 5.5 & 60.3 & 75.1 & 55.0 & 25.9 & 71.9 & 76.2 & 15.0 \\
\quad Claude Code Sonnet 5.5 & 47.3 & 63.8 & 37.5 & 25.4 & 47.0 & 64.9 & 19.1 \\
\quad Claude Code Haiku 5.5 & 42.7 & 45.5 & 35.1 & 25.3 & 55.2 & 63.7 & 7.2 \\
\quad Muse Code Ember 1.0 & 49.4 & 66.6 & 35.8 & 21.1 & 58.7 & 63.9 & 12.7 \\
\quad Gemini CLI Argon & 47.0 & 29.3 & 59.1 & 25.9 & 60.0 & 80.0 & \underline{23.4} \\
\quad Gemini CLI Hydrogen & 59.1 & 63.5 & 45.4 & \underline{33.3} & 78.1 & 77.8 & \textbf{27.7} \\
\multicolumn{8}{@{}l}{\textit{Extraction APIs}} \\
\quad LlamaExtract Agentic Plus Max Effort & \underline{66.2} & 84.0 & 60.1 & 32.8 & \underline{80.3} & 85.4 & 0.0 \\
\quad LlamaExtract Agentic Plus & 62.4 & 83.0 & 61.4 & 33.1 & 74.7 & 65.1 & 0.0 \\
\quad LlamaExtract Agentic & 61.4 & \underline{85.6} & 60.9 & 33.1 & 72.2 & 56.9 & 0.0 \\
\quad LlamaExtract Cost Effective & 49.4 & 55.6 & 46.8 & 25.4 & 62.4 & 70.8 & 0.0 \\
\quad Datalab Accurate & 46.5 & 69.9 & 49.6 & 16.8 & 55.9 & 32.3 & 0.0 \\
\quad Extend Operator & 57.7 & 80.3 & \textbf{78.2} & 18.3 & 61.2 & 48.9 & 0.0 \\
\quad Pulse Effort & 19.2 & 26.8 & 25.7 & 14.5 & 10.3 & 27.6 & 0.0 \\
\quad Reducto Deep & 45.3 & 49.6 & 40.3 & \textbf{39.3} & 59.5 & 53.2 & 0.0 \\
\bottomrule
\end{tabular}
\caption{Grounding F1 by challenge under the three-hour execution budget across the \proboxeddocs{} documents with bounding-box annotations. A value is grounded when it is correct and its box overlaps an accepted box at IoU $\ge 0.5$; the second header row gives the share of expected values with a box. Bold and underline mark the highest and second-highest scores in each column.}
\label{tab:grounding}
\end{table}

%% file: tables/timeouts.tex
\begin{table}[!htb]
\centering
\small
\setlength{\tabcolsep}{5pt}
\begin{tabular}{@{}lrrrrrr@{}}
\toprule
 & \multicolumn{3}{c}{\textbf{Minutes per document}} & \multicolumn{3}{c}{\textbf{Runs scoring 0}} \\
\cmidrule(lr){2-4}\cmidrule(lr){5-7}
\textbf{System} & \textbf{Mean} & \textbf{Median} & \textbf{Max} & \textbf{$>$3 hours} & \textbf{Failed} & \makecell{Finished\\zero F1} \\
\midrule
\multicolumn{7}{@{}l}{\textit{Extraction APIs}} \\
\quad LlamaExtract Agentic Plus Max Effort & 6.9 & 5.3 & 20 & 0 & 0 & 0 \\
\quad LlamaExtract Agentic Plus & 3.5 & 2.9 & 21 & 0 & 0 & 0 \\
\quad LlamaExtract Agentic & 3.2 & 2.7 & 11 & 0 & 0 & 0 \\
\quad LlamaExtract Cost Effective & 5.6 & 5.4 & 13 & 0 & 0 & 8 \\
\quad Datalab Accurate & 8.6 & 6.0 & 46 & 0 & 0 & 7 \\
\quad Extend Operator & 5.3 & 3.9 & 22 & 0 & 0 & 7 \\
\quad Pulse Effort & 8.2 & 6.8 & 25 & 0 & 1 & 1 \\
\quad Reducto Deep & 5.1 & 5.2 & 9 & 0 & 0 & 5 \\
\multicolumn{7}{@{}l}{\textit{Coding agents}} \\
\quad Codex GPT-6 Astra & 9.3 & 7.4 & 31 & 0 & 0 & 0 \\
\quad Codex GPT-6.1 Sol & 6.8 & 3.6 & 46 & 0 & 0 & 1 \\
\quad Codex GPT-6 Luna & 14.0 & 11.9 & 49 & 0 & 0 & 2 \\
\quad Codex GPT-6.1 Terra & 4.3 & 3.4 & 14 & 0 & 0 & 0 \\
\quad Claude Code Opus 5.5 & 17.2 & 11.8 & 59 & 0 & 0 & 0 \\
\quad Claude Code Sonnet 5.5 & 14.9 & 9.8 & 57 & 0 & 0 & 1 \\
\quad Claude Code Haiku 5.5 & 15.5 & 12.3 & 62 & 0 & 17 & 0 \\
\quad Muse Code Ember 1.0 & 15.3 & 11.2 & 60 & 0 & 0 & 1 \\
\quad Gemini CLI Argon & 28.7 & 23.4 & 154 & 0 & 30 & 0 \\
\quad Gemini CLI Hydrogen & 20.9 & 15.9 & 75 & 0 & 0 & 12 \\
\midrule
\textbf{All} & & & & 0 & 48 & 45 \\
\bottomrule
\end{tabular}
\caption{Wall-clock time per document over runs with a recorded duration, and runs scoring zero: runs over three hours, execution failures, and completed runs with zero Value F1. }
\label{tab:timeouts}
\end{table}

%% file: tables/results_overall_no_limit.tex
\begin{table}[!htb]
\centering
\small
\setlength{\tabcolsep}{4pt}
\begin{tabular}{@{}lrrrrrr@{}}
\toprule
 & \multicolumn{2}{c}{\textbf{Perfect Value F1 (\%)}} & \multicolumn{2}{c}{\textbf{Value F1}} & \multicolumn{2}{c}{\textbf{Grounding F1}} \\
\cmidrule(lr){2-3}\cmidrule(lr){4-5}\cmidrule(lr){6-7}
\textbf{System} & \textbf{30 min} & \textbf{3 hours} & \textbf{30 min} & \textbf{3 hours} & \textbf{30 min} & \textbf{3 hours} \\
\midrule
\multicolumn{7}{@{}l}{\textit{Extraction APIs}} \\
\quad LlamaExtract Agentic Plus Max Effort & 56 & 56 & 92.7 & 92.7 & 66.2 & 66.2 \\
\quad LlamaExtract Agentic Plus & 36 & 36 & 81.2 & 81.2 & 62.4 & 62.4 \\
\quad LlamaExtract Agentic & 38 & 38 & 80.6 & 80.6 & 61.4 & 61.4 \\
\quad LlamaExtract Cost Effective & 0 & 0 & 47.2 & 47.2 & 49.4 & 49.4 \\
\quad Datalab Accurate & 42 & 42 & 80.3 & \textbf{82.5} & 44.0 & \textbf{46.5} \\
\quad Extend Operator & 34 & 34 & 73.9 & 73.9 & 57.7 & 57.7 \\
\quad Pulse Effort & 19 & 19 & 72.1 & 72.1 & 19.2 & 19.2 \\
\quad Reducto Deep & 1 & 1 & 58.4 & 58.4 & 45.3 & 45.3 \\
\multicolumn{7}{@{}l}{\textit{Coding agents}} \\
\quad Codex GPT-6 Astra & 59 & 59 & 95.3 & \textbf{96.3} & 71.0 & \textbf{72.2} \\
\quad Codex GPT-6.1 Sol & 13 & 13 & 74.8 & \textbf{75.8} & 64.0 & \textbf{65.0} \\
\quad Codex GPT-6 Luna & 0 & 0 & 48.8 & \textbf{53.1} & 44.3 & \textbf{47.6} \\
\quad Codex GPT-6.1 Terra & 27 & 27 & 74.2 & 74.2 & 60.2 & 60.2 \\
\quad Claude Code Opus 5.5 & 51 & \textbf{61} & 75.4 & \textbf{93.1} & 45.3 & \textbf{60.3} \\
\quad Claude Code Sonnet 5.5 & 33 & \textbf{34} & 74.9 & \textbf{86.8} & 40.6 & \textbf{47.3} \\
\quad Claude Code Haiku 5.5 & 11 & \textbf{12} & 50.6 & \textbf{60.2} & 37.7 & \textbf{42.7} \\
\quad Muse Code Ember 1.0 & 35 & \textbf{37} & 73.4 & \textbf{81.2} & 43.1 & \textbf{49.4} \\
\quad Gemini CLI Argon & 29 & \textbf{40} & 42.5 & \textbf{61.8} & 32.4 & \textbf{47.0} \\
\quad Gemini CLI Hydrogen & 42 & \textbf{46} & 63.9 & \textbf{80.3} & 43.9 & \textbf{59.1} \\
\bottomrule
\end{tabular}
\caption{Scores with a 30-minute limit and the primary three-hour limit, on the same saved outputs. \textbf{Bold} marks a score that the limit changes.}
\label{tab:results-overall-no-limit}
\end{table}

%% file: tables/cost_by_challenge.tex
\begin{table}[!htb]
\centering
\small
\setlength{\tabcolsep}{4pt}
\begin{tabular}{@{}lrrrrrr@{}}
\toprule
\textbf{System} & \textbf{Interp.} & \textbf{Terms} & \textbf{Aggr.} & \textbf{Diagr.} & \textbf{Charts/maps} & \textbf{Long} \\
\midrule
\multicolumn{7}{@{}l}{\textit{Extraction APIs}} \\
\quad LlamaExtract Agentic Plus Max Effort & 10.8 & 8.3 & 8.0 & 19.9 & 19.5 & 332.3 \\
\quad LlamaExtract Agentic Plus & 7.5 & 7.5 & 7.5 & 7.5 & 7.5 & 7.5 \\
\quad LlamaExtract Agentic & 3.1 & 3.1 & 3.1 & 3.1 & 3.1 & 3.1 \\
\quad LlamaExtract Cost Effective & 1.0 & 1.0 & 1.0 & 1.0 & 1.0 & 1.0 \\
\quad Datalab Accurate & 3.4 & 2.6 & 3.9 & 8.1 & 5.2 & 40.5 \\
\quad Extend Operator & 10.0 & 8.8 & 8.8 & 12.5 & 9.2 & 8.8 \\
\quad Pulse Effort & 10.5 & 10.5 & 10.5 & 10.5 & 10.5 & 10.5 \\
\quad Reducto Deep & 6.3 & 5.6 & 5.8 & 6.0 & 6.5 & 7.0 \\
\multicolumn{7}{@{}l}{\textit{Coding agents}} \\
\quad Codex GPT-6 Astra & 17.0 & 14.2 & 17.6 & 19.3 & 15.1 & 1139.0 \\
\quad Codex GPT-6.1 Sol & 2.5 & 1.6 & 1.4 & 2.2 & 1.5 & 173.0 \\
\quad Codex GPT-6 Luna & 0.3 & 0.3 & 0.4 & 0.4 & 0.5 & 11.1 \\
\quad Codex GPT-6.1 Terra & 0.9 & 1.1 & 1.0 & 1.4 & 1.0 & 43.7 \\
\quad Claude Code Opus 5.5 & 9.4 & 7.0 & 5.7 & 12.7 & 7.8 & 466.5 \\
\quad Claude Code Sonnet 5.5 & 4.3 & 3.9 & 2.7 & 6.1 & 4.8 & 218.8 \\
\quad Claude Code Haiku 5.5 & 2.5 & 2.0 & 1.3 & 1.1 & 1.3 & 45.5 \\
\quad Muse Code Ember 1.0 & 0.8 & 0.7 & 0.7 & 1.9 & 1.4 & 65.0 \\
\quad Gemini CLI Argon & 15.5 & 9.1 & 10.4 & 13.0 & 22.5 & 314.1 \\
\quad Gemini CLI Hydrogen & 7.2 & 5.5 & 5.3 & 9.2 & 8.0 & 466.4 \\
\bottomrule
\end{tabular}
\caption{Mean document-level cost per page (\textcent) by challenge. A long chart sheet is one page, so its column is the mean cost per sheet. Abbreviations: Interp.\ (Interpretation rules), Terms (Custom terminology), Aggr.\ (Conditional aggregation), Diagr.\ (Relational diagrams), Charts and maps, and Long (Long single pages).}
\label{tab:cost-by-challenge}
\end{table}

%% file: tables/hardest_documents.tex
\begin{table}[!htb]
\centering
\small
\setlength{\tabcolsep}{1.8pt}
\renewcommand{\arraystretch}{0.95}
\begin{tabularx}{\textwidth}{@{}>{\raggedright\arraybackslash}X>{\raggedright\arraybackslash}p{5.5em}>{\centering\arraybackslash}p{1.65em}>{\centering\arraybackslash}p{1.65em}>{\centering\arraybackslash}p{1.65em}>{\centering\arraybackslash}p{1.65em}>{\centering\arraybackslash}p{1.65em}>{\centering\arraybackslash}p{1.65em}>{\centering\arraybackslash}p{1.65em}>{\centering\arraybackslash}p{1.65em}>{\centering\arraybackslash}p{1.65em}>{\centering\arraybackslash}p{1.65em}>{\centering\arraybackslash}p{1.65em}>{\centering\arraybackslash}p{1.65em}>{\centering\arraybackslash}p{1.65em}>{\centering\arraybackslash}p{1.65em}>{\centering\arraybackslash}p{1.65em}>{\centering\arraybackslash}p{1.65em}>{\centering\arraybackslash}p{1.65em}>{\centering\arraybackslash}p{1.65em}r@{}}
\toprule
 & & \multicolumn{8}{c}{\textbf{Extraction APIs}} & \multicolumn{10}{c}{\textbf{Coding agents}} & \\
\cmidrule(lr){3-10}\cmidrule(lr){11-20}
\textbf{Document} & \textbf{Challenge} & \rotatebox{90}{\textbf{LlamaExtract Agentic Plus Max Effort}} & \rotatebox{90}{\textbf{LlamaExtract Agentic Plus}} & \rotatebox{90}{\textbf{LlamaExtract Agentic}} & \rotatebox{90}{\textbf{LlamaExtract Cost Effective}} & \rotatebox{90}{\textbf{Datalab Accurate}} & \rotatebox{90}{\textbf{Extend Operator}} & \rotatebox{90}{\textbf{Pulse Effort}} & \rotatebox{90}{\textbf{Reducto Deep}} & \rotatebox{90}{\textbf{Codex GPT-6 Astra}} & \rotatebox{90}{\textbf{Codex GPT-6.1 Sol}} & \rotatebox{90}{\textbf{Codex GPT-6 Luna}} & \rotatebox{90}{\textbf{Codex GPT-6.1 Terra}} & \rotatebox{90}{\textbf{Claude Code Opus 5.5}} & \rotatebox{90}{\textbf{Claude Code Sonnet 5.5}} & \rotatebox{90}{\textbf{Claude Code Haiku 5.5}} & \rotatebox{90}{\textbf{Muse Code Ember 1.0}} & \rotatebox{90}{\textbf{Gemini CLI Argon}} & \rotatebox{90}{\textbf{Gemini CLI Hydrogen}} & \textbf{Mean} \\
\midrule
Water loss audit 02 & Diagrams & 50 & 20 & 25 & 17 & 59 & 14 & 24 & 18 & 76 & \textbf{100} & 15 & 25 & \textbf{100} & 14 & 18 & 23 & 0 & 0 & 33 \\
Time location chart 02 & Long pages & 57 & 78 & 2 & 1 & 0 & 0 & 17 & 0 & \textbf{99} & 96 & 0 & 34 & \textbf{99} & 0 & 0 & 82 & 0 & \textbf{99} & 37 \\
TV post buy affidavit 01 & Aggregation & 17 & 37 & 37 & 43 & 40 & 17 & 37 & 37 & \textbf{100} & 34 & 37 & 17 & 37 & 66 & 40 & 17 & 40 & 17 & 37 \\
Time location chart 01 & Long pages & 99 & 62 & 0 & 0 & 0 & 0 & 0 & 0 & \textbf{100} & 86 & 0 & 9 & 56 & 98 & 0 & 81 & 0 & 96 & 38 \\
Wireline log 01 & Long pages & 98 & 7 & 2 & 0 & 0 & 0 & 9 & 14 & \textbf{99} & 90 & 12 & 6 & 95 & 96 & 36 & 65 & 28 & 99 & 42 \\
HVAC balancing report 01 & Diagrams & 85 & 22 & 91 & 30 & 53 & 19 & 27 & 21 & 91 & 72 & 15 & 18 & \textbf{100} & \textbf{100} & 0 & 25 & 0 & 0 & 43 \\
Wireline log 02 & Long pages & 99 & 11 & 16 & 0 & 0 & 0 & 16 & 0 & \textbf{99} & 89 & 0 & 3 & 97 & 87 & 0 & 66 & 98 & 99 & 43 \\
Container demurrage claim 01 & Aggregation & 38 & 35 & \textbf{100} & 35 & \textbf{100} & \textbf{100} & 32 & 38 & 38 & 35 & 29 & 38 & 18 & 18 & 35 & 32 & 0 & \textbf{100} & 46 \\
Insurance programme file 01 & Diagrams & \textbf{96} & 44 & 40 & 19 & 42 & 52 & 33 & 33 & \textbf{96} & 33 & 38 & 33 & \textbf{96} & \textbf{96} & 19 & 33 & 0 & 33 & 46 \\
Probate court file 01 & Diagrams & \textbf{100} & 57 & 34 & 30 & 38 & 35 & 75 & 27 & \textbf{100} & 34 & 33 & 34 & \textbf{100} & 38 & 8 & 30 & 0 & 63 & 47 \\
\bottomrule
\end{tabularx}
\caption{Value F1 (\%) on the ten documents with the lowest mean Value F1 across the \proPerDocSystems{} systems, under the three-hour budget, lowest first. Scores are rounded to whole numbers, except that a cell reads 100 only when the document is perfect. Failed runs score 0.}
\label{tab:hardest}
\end{table}

%% file: sections/appendix/failures.tex
\FloatBarrier
\subsection{Case Selection}

The qualitative case review focuses on documents for which completed outputs disagree.
For each challenge, we originally selected the three documents with the largest Value F1 spread among finished runs in the initial twelve-system snapshot.
For the examples below, we inspect the outputs selected for the reported evaluation, comparing returned values with the answer key and, for coding agents, inspecting the final message.
They distinguish what was returned from what the error pattern suggests.

\subsection{Examples}

\paragraph{Interpretation rules.}
The reviewed outputs omit records.
On the patient record, Luna returned 91 of 1,277 result rows and stated that it had included only the first weekly report.
On the investor report, Reducto returned only one or two revenue rows per company where the document contains up to 52.

\paragraph{Custom terminology.}
Several errors occur on fields whose meanings are defined locally.
On the import-duty reconciliation, one reviewed Codex output matched the ordinary reading rather than the packet's definition, while Sol returned null for every duty and refund field.
On the spread-betting account, Luna stated that the settlement schedule and definitions appendix were absent and left the computed fields null.
Sol's race points were wrong by exactly one on 29 of 58 lines, consistent with substituting decimal odds for odds to one.
On the subscription-seat review, Reducto classified three unpaid order forms as paid.
These patterns locate the errors without proving which definitions a system read.

\paragraph{Conditional aggregation.}
Incorrect totals reflect both rule application and transcription problems.
In the election packet, short ballots belong in the district total but not the valid-ballot count.
Sol assigned between 1,816 and 2,573 additional ballots to the valid category in every district.
On the library-aid claim, Luna matched neither the defined count nor the simplified count.
It also misread a month code in the scan.
On the engine-utilization return, Sol overcounted cycles in every month, consistent with missing exclusions.

\paragraph{Relational diagrams.}
The reviewed outputs leave diagram-dependent fields unanswered.
On the HVAC balancing report, Luna left all 72 per-floor values null and stated that it could not resolve the totals from the diagram.
On the fiber-splice record, Extend listed every circuit but returned null for splice counts, losses, and lengths.

\paragraph{Charts and maps.}
The reviewed chart cases contain incorrect readings or selections, but their causes are not always identifiable.
On the tower-coverage credit, Sol's zone counts disagree with the expected counts from the coverage map.
These patterns suggest chart or map-reading errors but were not all verified through page inspection.

\paragraph{Long single pages.}
Empty outputs can arise from both provider restrictions and unsuccessful extraction.
Datalab rejected two reviewed sheets because their width exceeded 7,200 pixels and returned null for every value on the third, while Extend completed all seven sheets without a reported execution error but returned only null values.
These outcomes all receive zero, but a provider rejection is distinct from an incorrect reading of an accepted input.

%% file: sections/appendix/additional_analysis.tex
\input{sections/additional_analysis_macros}

\subsection{Document-Level Score Distributions}
\label{app:score-distributions}

\begin{figure}[!htb]
\centering
\includegraphics{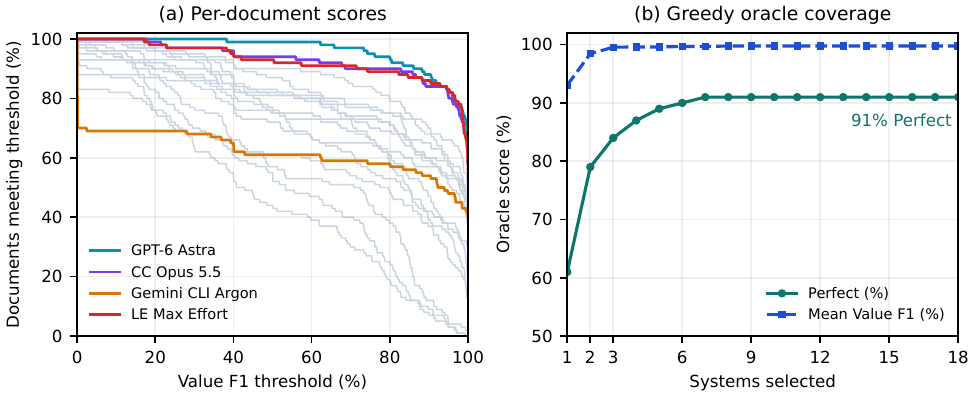}
\caption{\textbf{Left:} Percentage of documents meeting each Value F1 threshold; the remaining systems appear in gray. \textbf{Right:} Oracle scores as systems are added greedily to maximize additional perfect-document coverage, breaking ties by oracle mean Value F1. Both panels use all 100 documents under the three-hour budget.}
\label{fig:distributions-oracle}
\end{figure}

For several strong extractors, per-document Value F1 scores cluster near 100\%, with a long left tail of low-scoring documents (\Cref{fig:distributions-oracle}, left). For example, Gemini CLI Argon has a median Value F1 of \analysisArgonMedian\% but a mean of \analysisArgonMean\%.

\subsection{System Complementarity and Oracle Coverage}
\label{app:complementarity}

\begin{figure}[!htb]
\centering
\includegraphics{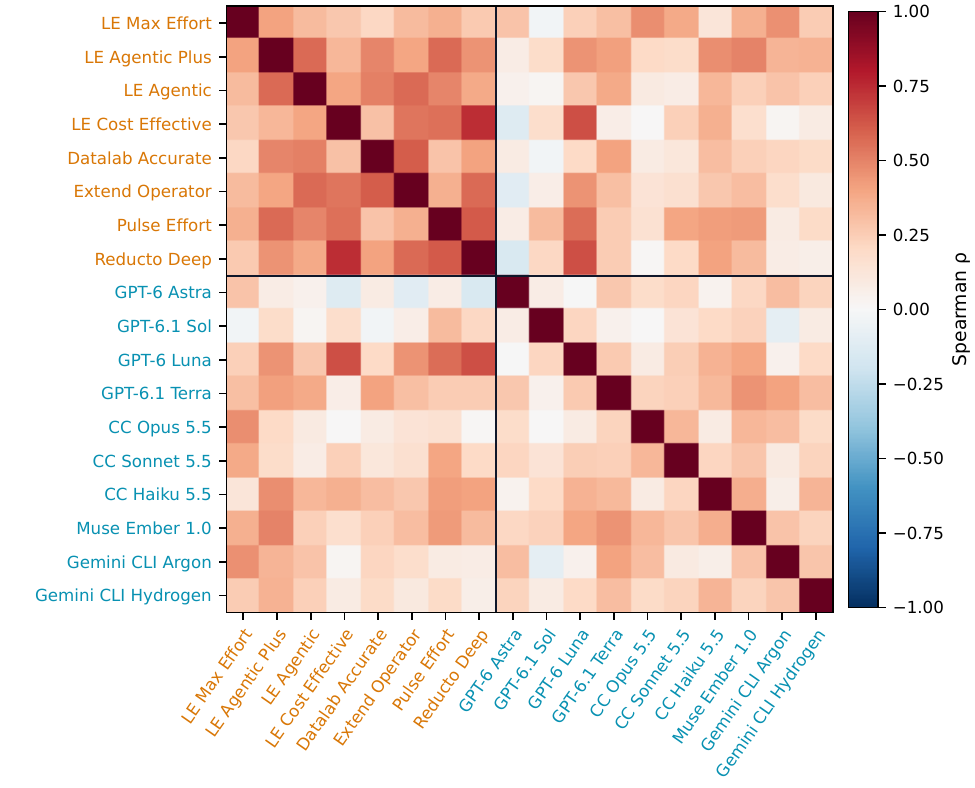}
\caption{Spearman correlation of per-document Value F1 across all \pronsystems{} systems. Orange labels identify extraction APIs; teal labels identify coding agents. Lines separate the two groups.}
\label{fig:score-agreement}
\end{figure}

\Cref{fig:score-agreement} compares per-document Value F1 across systems, grouped by extractor type. We consider an oracle that routes each document to the best-performing system in a selected set. Adding systems greedily increases perfect-document coverage from \analysisOracleOne\% with one system to \analysisOracleTwo\% with two and \analysisOracleThree\% with three (\Cref{fig:distributions-oracle}, right). Across all \pronsystems{} systems, at least one extractor is perfect on \analysisOracleAll\% of documents.

\subsection{Reasoning-Effort Scaling}
\label{app:effort-scaling}

We compare Low, Medium, and Extra High effort on all 100 documents using a one-hour execution budget (\Cref{tab:effort-scaling}). Raising effort from Low to Extra High improves Value F1 for all three models (one-sided paired Wilcoxon, $p < 0.001$), while increasing known spending and median latency. Even at Extra High, no configuration exceeds \analysisEffortBestPerfect\% perfect documents.

\input{tables/stat_effort}

%% file: sections/additional_analysis_macros.tex
\newcommand{\analysisArgonMean}{61.8}
\newcommand{\analysisArgonMedian}{93.1}
\newcommand{\analysisOracleOne}{61}
\newcommand{\analysisOracleTwo}{79}
\newcommand{\analysisOracleThree}{84}
\newcommand{\analysisOracleAll}{91}
\newcommand{\analysisEffortBestPerfect}{64}
\newcommand{\analysisSolLowPerfect}{13}
\newcommand{\analysisSolHighPerfect}{48}
\newcommand{\analysisTerraLowPerfect}{11}
\newcommand{\analysisTerraHighPerfect}{55}
\newcommand{\analysisSonnetLowPerfect}{32}
\newcommand{\analysisSonnetHighPerfect}{64}

%% file: tables/stat_effort.tex
\begin{table}[!htb]
\centering
\small
\setlength{\tabcolsep}{5pt}
\begin{tabular}{@{}llrrrr@{}}
\toprule
System & Effort & Perfect (\%) & Value F1 [95\% CI] & Cost (\textcent/page) & Median (min) \\
\midrule
Codex GPT-6.1 Sol & Low & 13 & 75.8 [70.8, 80.7] & 13.9 & 3.6 \\
 & Medium & 28 & 88.2 [84.3, 91.7] & 18.9 & 5.2 \\
 & Extra High & 48 & 92.7 [89.4, 95.5] & 19.5 & 8.8 \\
\addlinespace[2pt]
Codex GPT-6.1 Terra & Low & 11 & 51.3 [44.7, 58.2] & 2.6 & 1.8 \\
 & Medium & 27 & 74.2 [68.2, 80.0] & 4.1 & 3.4 \\
 & Extra High & 55 & 89.3 [85.0, 93.1] & 12.3 & 8.1 \\
\addlinespace[2pt]
CC Sonnet 5.5 & Low & 32 & 78.2 [71.8, 84.3] & 5.5 & 6.9 \\
 & Medium & 33 & 84.8 [79.4, 89.7] & 20.0 & 9.9 \\
 & Extra High & 64 & 92.4 [88.2, 96.0] & 43.0 & 15.2 \\
\bottomrule
\end{tabular}
\caption{Reasoning effort on all 100 documents, with a one-hour execution budget per document. Failures score zero. Value F1 intervals use 10,000 paired document-bootstrap resamples. Cost is the mean document-level cost per page.}
\label{tab:effort-scaling}
\end{table}